\documentclass[journal]{IEEEtran}

\IEEEoverridecommandlockouts                              
\usepackage{amssymb}
\usepackage{amsmath} 
\usepackage{xcolor}  
\usepackage{multirow}
\usepackage{hyperref}
\usepackage{cite}
\usepackage{amsmath,graphicx,array}

\title{\LARGE \bf
Multi-modal Representation Fusion for Long-term  Microsurgical Anastomosis Trajectory Anticipation
}

\author{Yuxuan Liu, Yating Luo, Yunfei Luan, Xinyao Zhou,  Yuyang Zhuge,  Yao Guo, and Guang-Zhong Yang%
}

\begin{document}

\maketitle
\thispagestyle{empty}
\pagestyle{empty}

\begin{abstract}
Microsurgical anastomosis is one of the most challenging and sophisticated procedures for surgical robotics, requiring accurate manipulation of surgical instruments within a limited space to achieve precise operations. Accurate surgical trajectory anticipation is essential for surgical scene understanding, intraoperative planning, robotic manipulation, and reliable behaviour analysis. However, existing approaches mainly focus on features extracted from surgical videos and historical trajectories, while overlooking rich semantic and motion information that can provide more discriminative multi-modal representations for complex surgical instrument-tissue interactions. In this work, we propose a cross-modal feature fusion framework for long-term 2D and 3D microsurgical anastomosis trajectory anticipation. The proposed framework consists of a multi-modal representation extractor and a cross-modal feature fusion network. Semantic recognition of surgical objects and consecutive motion estimation are incorporated as multi-modal visual cues, providing complementary spatial and temporal guidance for instrument-tissue interaction characterization. A coarse-to-fine progressive fusion strategy with a long-short-term memory mechanism is introduced to capture temporal dependencies step by step. The training and evaluation are conducted on our established robotic platform and a comprehensive dataset containing both phantom and ex-vivo surgical scenarios is constructed. Extensive quantitative and qualitative experiments demonstrate that the proposed framework consistently outperforms existing methods and exhibits improved robustness across different surgical scenes, highlighting its potential for both intelligent robot-assisted and autonomous surgical systems.

\end{abstract}

\section{Introduction}
 \begin{figure}[!htbp]
    \centering
    \includegraphics[width=1.0\linewidth]{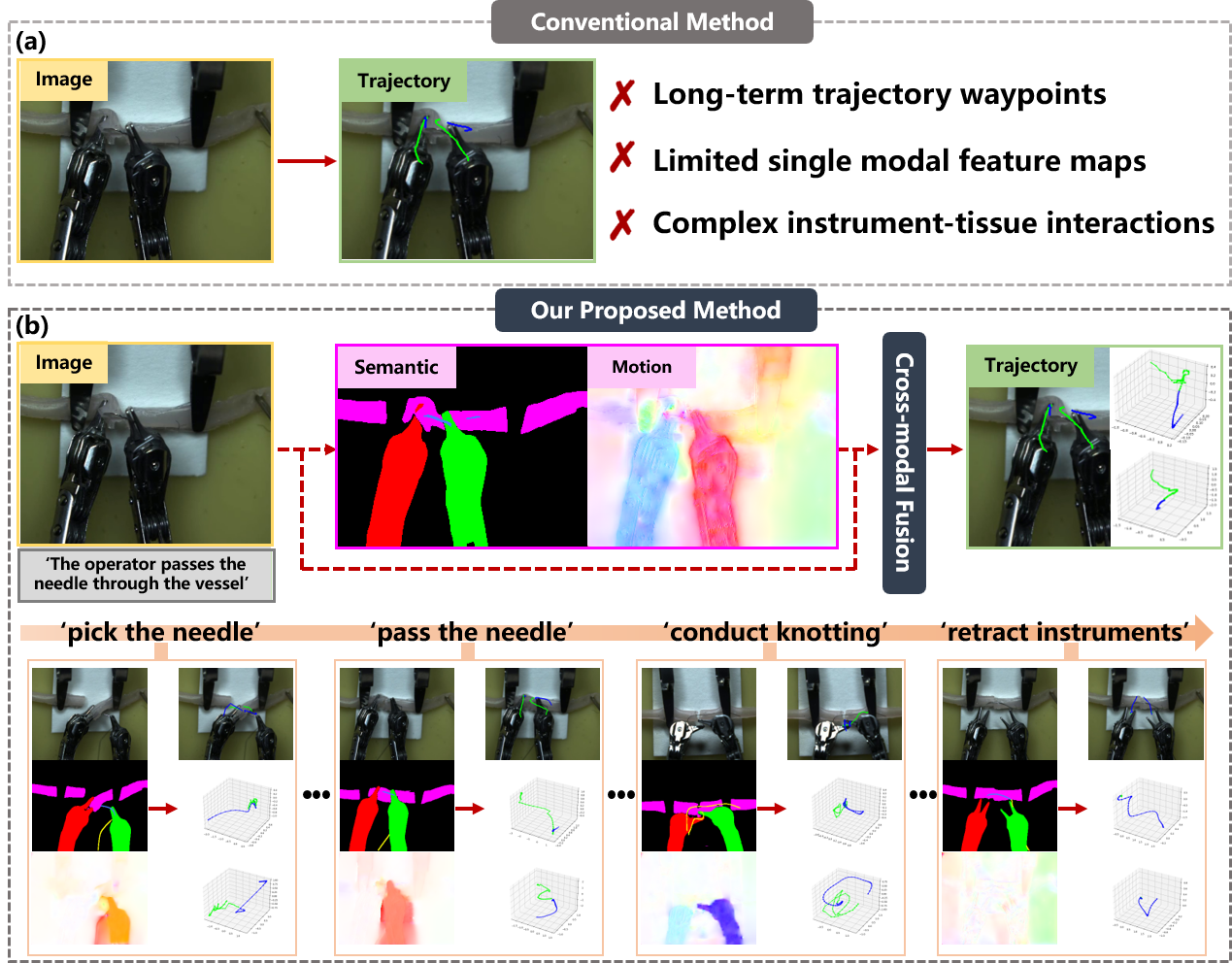}
    \caption{Illustration of the microsurgical anastomosis trajectory anticipation. (a) Existing methods perform this task directly based on either past trajectories or observed surgical videos, while overlooking diverse multi-modal representations for long-term prediction and complex instrument-tissue interactions. (b) Our proposed framework incorporates the fusion of multi-modal representations for long-term 2D and 3D trajectory anticipation on microsurgical anastomosis.}
    \label{fig:intro}
    \vspace{-10pt}
\end{figure}

Microsurgical robotics has experienced transformative progress in the field of minimally invasive surgery, providing surgeons with enhanced visual perception and robotic control within the submillimeter scale by introducing advanced robotic platforms and high-resolution multi-modal imaging techniques \cite{yang2018grand,dupont2025medical,kim2025srt}. Recent research efforts have concentrated on the development of reliable and precise robotic systems, leading to successful applications in diverse tasks such as flexible neural electrode implantation \cite{luo2026hierarchical,an2024microscopic} and retinal vein injection \cite{bian2026autonomous,zhang2025deep}. Among various microsurgical procedures, anastomosis is considered to be one of the most technically demanding operations and is extensively employed in clinical scenarios \cite{saeidi2022autonomous,haworth2026suturebot,huaulme2021micro}. 

Microsurgical anastomosis involves accurately reconstructing broken vessels with well-designed instruments and delicate operations of suturing threads and needles \cite{liu2026cross,choi2025surgmllmbench,zhou2026wrist}. By leveraging robotic assistance, microsurgical anastomosis can be conducted with submillimeter accuracy, improving the efficiency and autonomy level for broader clinical deployment. To achieve reliable autonomous microsurgical anastomosis, the ability to anticipate future motions and interactions between instruments and tissues is critical for intraoperative planning, surgical scene understanding, and policy generation \cite{wang2025idpoe,pei2025instrument,liu2025fpm}. Surgical trajectory anticipation aims to predict future movements of surgical tools in both 2D and 3D coordinate systems based on historical and current observations. Similar to the popular hand trajectory prediction for embodied artificial intelligence \cite{ma2025madiff,liu2022joint}, surgical trajectory anticipation plays the same role in surgical understanding, benefiting diverse downstream tasks for robot control and manipulation \cite{chen2025surgsora,biagini2025hierasurg}. 
Consequently, accurate trajectory anticipation is of great importance for instrument-tissue interaction characterization, intraoperative path planning, surgical behaviour understanding, and robot action generation, and has unleashed the power for both advancing more intelligent robot-assisted microsurgery and facilitating higher levels of anastomosis autonomy.

Recent progress in deep learning has provided opportunities for improved surgical scene understanding both in robot-assisted surgery and surgical autonomy \cite{kim2025srt,long2025surgical}. These advanced techniques have contributed to the development of surgical instrument recognition \cite{zhao2025rethinking}, motion estimation \cite{bian2023motion}, and 3D reconstruction \cite{liu2026simultaneous}, and extensive studies have focused on proposing a unified framework for multiple tasks simultaneously with multi-modal feature extraction \cite{ayobi2025pixel,li2025collaborative}. By incorporating unified feature maps, cross-modal fusion makes the network perform better on downstream tasks. Although surgical trajectory anticipation has attracted considerable attention recently in embodied surgical autonomy, existing approaches only use extracted visual features for prediction from collected surgical videos, along with observed past trajectories \cite{wang2025idpoe,shi2022ttm,mao2023led}, and ignore the rich visual contexts from surgical scene understanding, which hinders the accuracy and generalization ability for microsurgery.


As shown in Fig. \ref{fig:intro}, in this work, we propose a multi-modal feature representation fusion framework for long-term surgical trajectory prediction on microsurgical anastomosis and tackle several important challenges inherent to our framework. Firstly, existing works mainly focus on predicting the target region and affordance map instead of long-term trajectory waypoints, which can not comprehensively characterize the instrument-tissue interactions. Compared with the task that only considers the destination position, operating trajectories preserve the detailed interactions between surgical instruments and objects dynamically. To make the annotation of trajectories less labor-intensive, we introduce imitation learning on our established robotic platform under the demonstration of the experts. Secondly, directly regressing the surgical trajectory from videos hinders the accuracy and generalization, which results in the low confidence and poor interpretability of generated results. Complex environments in microsurgery, coupled with specularities, motion blurs, and out of focus, make the deep neural network difficult to extract distinctive visual features without priors. Consequently, additional auxiliary multi-modal feature representations are prerequisite for robust future anticipation. Thirdly, how to fuse multi-modal features of high-level surgical scene understanding is of great importance to improve the performance of prediction, especially for long-term sequences. By seamlessly integrating cross-modal information, the proposed framework has the potential to stably generalize to different scenarios and accurately characterize the instrument-tissue interactions for microsurgical anastomosis.

To address the aforementioned challenges, we attempt to anticipate long-term surgical trajectories both in 2D and 3D space, and our proposed framework consists of two separate modules, i.e., a multi-modal representation extraction network and a cross-modal feature fusion network. The proposed multi-modal feature extractor simultaneously obtains semantic segmentations and motion fields in a multi-task semi-supervised manner. The proposed cross-modal fusion network conducts feature fusion in a coarse-to-fine scheme based on a well-designed transformer-based architecture with a progressive fusion scheme, and the 2D and 3D surgical trajectories are predicted based on fused feature maps. To evaluate our proposed framework, we have established a surgical robotic platform for microsurgery anastomosis, and a large amount of surgical data has been collected on different anastomosis scenes. By integrating our proposed method into our robotic platform, quantitative and qualitative results on both phantom and ex-vivo cases have demonstrated the superior performance and training strategy compared with existing methods, proving its potential for clinical microsurgery adoption.

In summary, the main contributions of this paper include:
\begin{itemize}
\item A multi-modal representation fusion framework is proposed for long-term microsurgical anastomosis 2D and 3D trajectory anticipation, which consists of a multi-modal representation extraction network and a cross-modal feature fusion network.
\item Auxiliary semantic and motion maps are introduced to characterize the feature representation of instrument-tissue interactions by a semi-supervised multi-modal feature extractor, and a progressive fusion scheme is designed to model the temporal dependencies of long-term feature sequences, which provides distinct visual cues for precise trajectory anticipation.
\item A large amount of microsurgical anastomosis data is collected to evaluate the proposed method on the phantom and ex-vivo cases based on the established surgical robotic platform.
\item Quantitative and qualitative experiments are conducted to demonstrate the effectiveness of the proposed framework for long-term 2D and 3D trajectories, which proves the potential applications on real-world microsurgical scenarios.
\end{itemize}

\begin{figure*}[!htbp]
    \centering
    \includegraphics[width=1.0\linewidth]{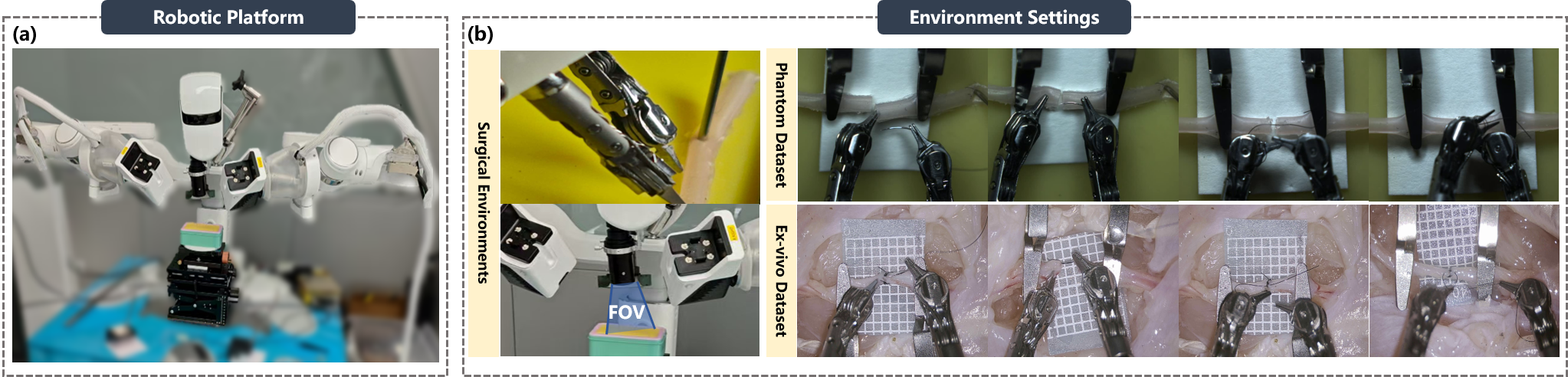}
    \caption{Overview of our established microsurgical anastomosis robotic platform and collected dataset. (a) Our established robot implements a dual-arm robotic system for anastomosis and a high-resolution microscope to capture surgical videos. (b) Our collected dataset includes surgical data for two different scenarios, i.e., anastomosis on the phantom vessel and ex-vivo vessel. }
    \label{fig:robot_system}
    \vspace{-10pt}
\end{figure*}

\section{Related Works}
\subsection{Multi-modal Representation Learning for Surgical Videos}
Recent advances in deep learning have significantly promoted surgical scene understanding by extracting rich visual representations from surgical videos \cite{carstens2025artificial,min2025innovating}. Early studies mainly focused on one single task, such as instrument segmentation, phase recognition, and depth prediction \cite{shen2023branch,czempiel2020tecno,bardozzo2022stasis}. Although these methods have achieved promising performance on specific applications, they do not emphasize the inherent relationships among different visual modalities. To overcome this limitation, multi-task and multi-modal representation learning has emerged as an effective technique for learning a unified feature representation. Existing studies have demonstrated the effectiveness of the joint optimization of instrument segmentation and depth estimation \cite{li2025collaborative,psychogyios2022msdesis}, depth estimation and motion prediction \cite{liu2019unsupervised,shao2022self,liu2026simultaneous}, image inpainting and depth estimation \cite{zhang2023self}, and phase recognition and segmentation \cite{ayobi2025pixel}. More recently, several works have further incorporated semi-supervised and self-supervised learning into the multi-task learning scheme, reducing the demand on annotation \cite{liu2026itcnet,wang2025improving}. These approaches indicate that heterogeneous visual cues, including semantics, temporal motion, and structural information, are highly complementary and can provide inherent discriminative representations for instrument-tissue interaction characterization and surgical scene understanding.

\subsection{Operating  Surgical Trajectory Anticipation}
Surgical trajectory anticipation has recently attracted increasing attention in embodied surgical autonomy. Existing methods commonly formulate trajectory anticipation as a time series prediction problem from either past observations or surgical videos \cite{chen2025real,wang2025idpoe,shi2022ttm,weerasinghe2024multimodal}. Previous studies implemented long-short-term memory networks to model temporal dependencies from past trajectories \cite{alahi2016social,xie2020motion}, and recent studies have increasingly adopted transformer-based architectures to capture long-range temporal relationships by attention mechanisms \cite{bao2023usst,shi2022ttm,mao2023led}. More recently, diffusion-based approaches have been designed to generate future trajectories and achieve better performance due to the strong capability of denoising procedure \cite{ma2025madiff,wang2025idpoe}. Meanwhile, several studies in hand trajectory prediction have demonstrated that introducing additional representations, including language instructions, motion cues, and semantic guidance can facilitate prediction \cite{choi2026egghand,ma2025madiff,ma2025novel}. Such auxiliary information provides high-level contextual cues that can improve the stability and consistency of future trajectories. However, these multi-modal representations have not been fully explored in surgical scenarios, and existing approaches still predominantly rely on visual features of RGB videos. Consequently, the accuracy and generalization ability are hindered in microsurgical anastomosis, resulting from a limited operating space with tiny instruments and complex instrument-tissue interactions.

\section{Surgical Robotic System and Data Collection}
\subsection{System Setup}
To demonstrate the effectiveness of our proposed framework for microsurgical anastomosis, we have established a robot platform for data collection and evaluation. The overview of our surgical robot is shown in Fig. \ref{fig:robot_system}-(a), which consists of three fundamental parts. The first part is the vision system with a high-resolution microscope to provide the top-down-view intraoperative perception. The second part is the dual-arm robotic system equipped with instruments for surgical operations. Our robot system is built upon two Physik Instrumente 6-axis micro-motors with a repeat accuracy of 0.06$\mu m$, which guarantees the accuracy of robot control at the submillimeter scale. The two instruments on the left and right hands are used to manipulate objects, like suturing needles and threads, precisely and stably. The third part is the surgical environmental settings where vessels with different scales are incorporated for anastomosis, and both phantom and ex-vivo scenarios are included in our data collection.

Following our previous work \cite{liu2026cross}, the entire procedure of microsurgical anastomosis includes two main parts, i.e., suture making and knotting. During suture making, the left instrument is used to fix the vessel, and the right instrument is used to hold the needle and pass it through the vessel. During knotting, the right instrument grabs the thread and performs circular rotations twice to tie a knot. The entire process can be divided into ten specific surgical phases in terms of different actions conducted by the left and right instruments.

\subsection{Data collection and preprocessing}
To improve the diversity of our collected data, we have asked experienced operators to perform anastomosis in an expert-demonstration manner. All operators control our established robot in a teleoperation scheme and are well-trained with practice for more than one week to ensure the time for either suture making or knotting is within five minutes. Videos are captured by the high-resolution microscope and synchronized with the derived kinematic data from the robot motor.  The derived kinematic data contains the state and position of the instruments and is processed to derive the 3D trajectories. In total, we have collected about 76k and 51k frames for our proposed method on the phantom case and ex-vivo case, respectively. The illustration and some examples of our data collection are shown in Fig. \ref{fig:robot_system}-(b).

To train our proposed framework in a semi-supervised learning manner, a small number of annotations are required. The annotation process is conducted by experts to obtain semantic segmentation maps for instruments, threads, needles, and operated vessels. To generate the instructions of surgical operations, annotations of the surgical phase and action are provided for each surgical sequence with ten predefined classes.  Surgical 2D trajectories are derived by a pretrained detection network \cite{tian2026yolov12} with manually annotated frames of 2D trajectory keypoints, and 3D trajectories are calculated based on kinematics data of the robot system and normalized to the canonical coordinate system. The overview of the data processing pipeline is illustrated in Fig.  \ref{fig:robot_system}-(c).

\section{Multi-modal Feature Representation Extraction for Instrument-Tissue Interaction Characterization}
For accurate surgical trajectory anticipation, visual feature extraction plays an important role in understanding complicated surgical operations and instrument-tissue interactions. Instead of directly predicting future sequences from videos, we attempt to extract multi-modal feature representations, including both semantic and motion cues.  In this section, we introduce the proposed multi-modal representation extractor in a semi-supervised multi-task learning scheme. Our proposed network and derived multi-modal visual cues make it easier and more efficient for understanding anastomosis operations, and serve as auxiliary feature maps for long-term trajectory anticipation. Inspired by \cite{liu2026itcnet}, we have selected two distinctive representations, i.e., semantic segmentation and motion estimation. The semantic segmentation refers to the recognition of five different objects, i.e., left and right instruments, suture needle and thread, as well as the operated vessel. The temporal motion estimation refers to pixel-wise correspondence between consecutive frames. 

\begin{figure*}[!htbp]
    \centering
    \includegraphics[width=1.0\linewidth]{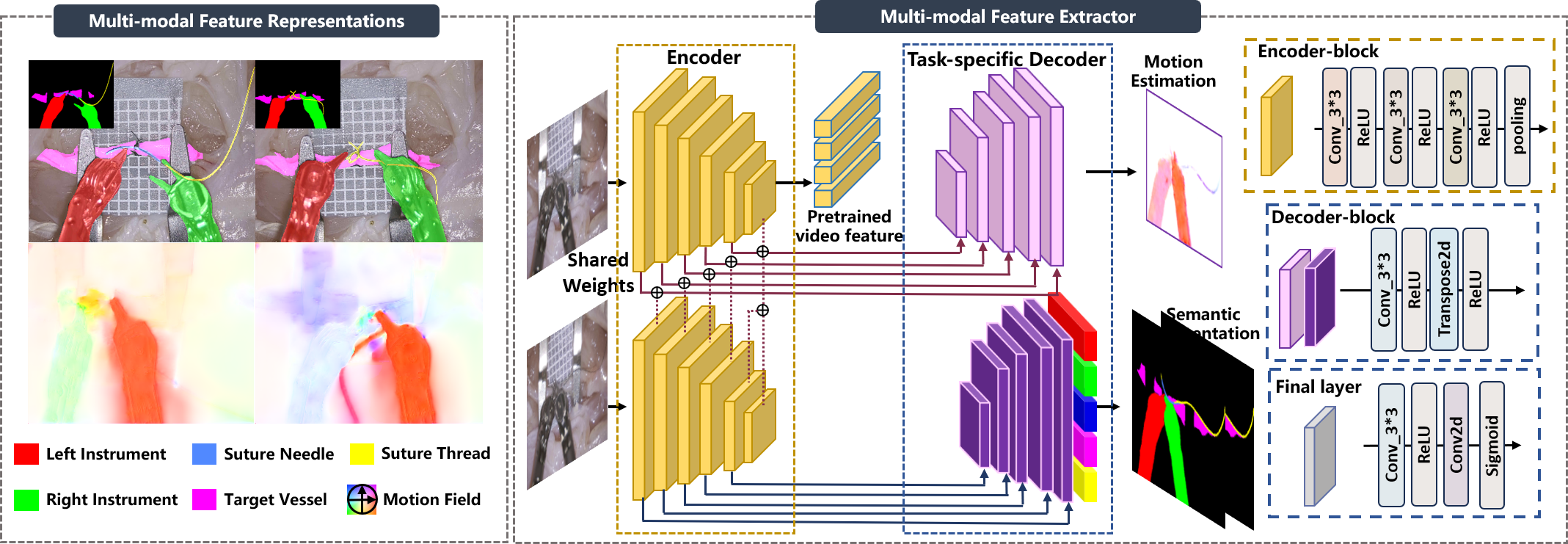}
    \caption{Overview of the proposed multi-modal feature representation extraction network. Five semantic segmentations and the motion field are predicted by the multi-modal feature extractor, where the left instrument, right instrument, needle, thread, and operated vessel are selected as the semantic guidance. The network introduces a multi-task learning scheme with a unified feature encoder and several task-specific decoders to extract different feature representations simultaneously.}
    \label{fig:visual_feature}
    \vspace{-10pt}
\end{figure*}

\subsection{Problem Definition}
Let $\mathrm{I}=\{I^t|1\leq t\leq T\}\in \mathbb{R}^{H \times W \times 3}$ represent the surgical video sequences captured by our vision system with $T$ images, where $H$ and $W$ refer to the height and width of each image, respectively. As shown in Fig. \ref{fig:visual_feature}, we propose to extract diverse representations for scene understanding and instrument-tissue interactions with our proposed multi-task feature extractor $\mathcal{F}_{\theta}(\cdot)$ with parameters $\theta$. We denote $\mathrm{Z}=\{\mathrm{M}^{l}, \mathrm{M}^r, \mathrm{M}^{nd}, \mathrm{M}^{tr}, \mathrm{M}^{vl},  \mathrm{F}\}$ for multi-modal visual representations, where $\mathrm{M} \in \mathbb{R}^{H \times W \times 1}$ represents the semantic segmentation of left instrument $\mathrm{M}^{l}$, right instrument $\mathrm{M}^{r}$, suture needle $\mathrm{M}^{nd}$, suture thread $\mathrm{M}^{tr}$, and target vessel $\mathrm{M}^{vl}$, and $\mathrm{F}  \in \mathbb{R}^{H \times W \times 2}$ represents the motion field of consecutive frames. Our multi-modal feature extraction network takes the surgical video as input and predicts the semantic segmentations and motion maps: $\{\mathrm{M}^{l}, \mathrm{M}^r, \mathrm{M}^{nd}, \mathrm{M}^{tr}, \mathrm{M}^{vl},  \mathrm{F}\} = \mathcal{F}(\mathrm{I})$. The training process of the multi-model representation extraction network can be defined as:
\begin{equation}
    \mathcal{F}_{\theta^*}=\arg \min_\theta \mathcal{L}_{feat}(\mathrm{I},\mathrm{Z},\mathrm{Z}^{gt}),
\end{equation}
where $\mathcal{L}_{feat}$ represents the loss function for multi-modal feature extraction and $\mathrm{Z}^{gt}$ represents the manual annotations.

\subsection{Network Architecture}
Following the commonly used network architecture for multi-task learning-based segmentation and motion estimation, our multi-modal feature extractor is built upon the residual UNet backbone. The proposed framework consists of a shared feature encoder and two task-specific decoders for simultaneously predicting semantic segmentations and motion fields. As shown in Fig. \ref{fig:visual_feature}, the basic block contains two group norm layers, two sigmoid functions, and two convolution layers with kernel $3\times3 $. The extracted unified feature maps are with the size of $\frac{H}{8}\times \frac{W}{8}$ in this work. Following the popular structure of UNet-based networks, we introduce the skip connections by concatenating the hierarchical unified feature maps $\{\frac{H}{2}\times \frac{W}{2}, \frac{H}{4}\times \frac{W}{4}, \frac{H}{8}\times \frac{W}{8}\}$ with task-specific decoders. The task-specific decoders have the same basic blocks as those in the shared feature encoder. The last convolution layer with kernel size $3\times 3$ is introduced to recover the corresponding feature channels. For the task of semantic segmentation of instruments, threads, needles, and target vessels, the output is converted into the probability map ranging from $0$ to $1$, and a threshold of 0.5 is selected to generate the binary masks. For motion prediction, the predicted motion fields are represented from $-1$ to $1$ as the pixel-wise displacement.  

\subsection{Network Training}
Previous works \cite{psychogyios2022msdesis,li2025collaborative} have demonstrated that multi-task learning of semantic segmentation and geometry estimation generates more robust shared feature representation and improves the performance of the individual task. In this work, we implement the multi-task learning scheme to extract diverse visual representations for surgical video understanding. 

The first step is to train the teacher network to generate smooth labels for semantic segmentation. Given a small amount of annotated data $\mathrm{M}^{gt}=\{\mathrm{M}^l_{gt},\mathrm{M}^r_{gt},\mathrm{M}^{nd}_{gt},\mathrm{M}^{tr}_{gt},\mathrm{M}^{vl}_{gt}\}$, we train several separate teacher networks $\{\mathcal{T}^{ins}(\cdot),\mathcal{T}^{nd}(\cdot),\mathcal{T}^{tr}(\cdot),\mathcal{T}^{vl}(\cdot)\}$ with the same architecture to perform semantic segmentation for instruments, suture needle, suture thread, and target vessel. Since severe occlusions frequently occur when anastomosis is operated on a microsurgical scale, one pixel may refer to different parts of semantic objects. Consequently, we define the semantic recognition into multiple binary segmentation instead of one multi-object segmentation task. The teacher network is trained based on Binary CrossEntropy Loss (BCE) $\mathcal{L}_{seg}=\mathrm{BCE}(\mathrm{M}_w,\mathrm{M}_{gt})$, where $\mathrm{M}_w$ represents the generated weak labels by teacher networks. The entire training process of different teacher networks can be summarized as:
\begin{equation}
\begin{cases}
\mathcal{T}^{ins}_{\theta^*}=\arg \min_{\theta} (\mathcal{L}_{seg}(\mathrm{M}^l_w,\mathrm{M}^l_{gt})+\mathcal{L}_{seg}(\mathrm{M}^r_w,\mathrm{M}^r_{gt}))
\\ 
\mathcal{T}^{nd}_{\theta^*}=\arg \min_{\theta} \mathcal{L}_{seg}(\mathrm{M}^{nd}_w,\mathrm{M}^{nd}_{gt})
\\
\mathcal{T}^{tr}_{\theta^*}=\arg \min_{\theta} \mathcal{L}_{seg}(\mathrm{M}^{tr}_w,\mathrm{M}^{tr}_{gt})
\\
\mathcal{T}^{vl}_{\theta^*}=\arg \min_{\theta} \mathcal{L}_{seg}(\mathrm{M}^{vl}_w,\mathrm{M}^{vl}_{gt})
\end{cases}.
\end{equation}

The second step is to conduct multi-task training for all visual representations with different task-specific decoders. The motion decoder predicts the motion field $\mathrm{F}$ and is constrained by the similarity term $\mathcal{L}_{mot}^s$ between transformed and target images and the regularization term $\mathcal{L}_{mot}^r$ of smoothness:
\begin{align}
&\mathcal{L}_{mot}^s=||I^{t_2}-\mathrm{F}^{t_{1,2}}\circ I^{t_1}||_1+\frac{(1-\mathrm{SSIM}(I^{t_2},\mathrm{F}^{t_{1,2}}\circ I^{t_1}))}{2}, \nonumber \\
&\mathcal{L}_{mot}^r = |\partial_x \mathrm{F}^{t_{1,2}}|e^{-||\partial_xI^{t_1}||} + |\partial_y \mathrm{F}^{t_{1,2}}|e^{-||\partial_yI^{t_1}||}.
\end{align}
The four segmentation decoders predict corresponding semantic masks for two instruments, suture needle, suture thread and target vessel under the supervision of pretrained teacher networks. In addition, we have added the cross-task consistency $\mathcal{L}_{ms}$ between the segmentation term and motion term, which consists of a motion-based geometric segmentation consistency $\mathcal{L}_{m\rightarrow s}$ and a segmentation-based edge-aware motion consistency $\mathcal{L}_{s\rightarrow m}$. The cross-task constraint facilitates the temporal cues between consistent motion fields and accurate instrument segmentation. In summary, the multitask training process can be defined as:
\begin{equation}
    \begin{cases}
        \mathcal{L}_{seg}= \alpha_{seg}^{ins}(\mathrm{BCE}(\mathrm{M}^l,\mathrm{M}^l_w) + \mathrm{BCE}(\mathrm{M}^r,\mathrm{M}^r_w)) \\ ~~~~~~~~~~+ \alpha_{seg}^{nd}\mathrm{BCE}(\mathrm{M}^{nd},\mathrm{M}^{nd}_w)  + \alpha_{seg}^{tr}\mathrm{BCE}(\mathrm{M}^{tr},\mathrm{M}^{tr}_w) \\ ~~~~~~~~~~+ \alpha_{seg}^{vl}\mathrm{BCE}(\mathrm{M}^{vl},\mathrm{M}^{vl}_w) \\
        \mathcal{L}_{mot} = \alpha_{mot}^{s}\mathcal{L}_{mot}^s + \alpha_{mot}^{r} \mathcal{L}_{mot}^r \\
        \mathcal{L}_{ms} = \mathcal{L}_{s\rightarrow m} +\mathcal{L}_{m \rightarrow s}\\
        \mathcal{F}_{\theta^*}=\arg \min_{\theta} \mathcal{L}_{feat}=\arg \min_{\theta}  (\mathcal{L}_{seg}+\mathcal{L}_{mot}+\mathcal{L}_{ms})
    \end{cases}
\end{equation}

To eliminate the influence of weak labels with poor quality, we separately finetune different decoders with annotated labels if available. During this procedure, the parameters of the shared feature encoder are frozen and the supervised training constraints are imposed on task-specific finetuning. For four segmentation decoders, we replace the generated weak labels with ground truth annotations if available and follow the training constraints mentioned above. During the inference stage, our well-designed framework can simultaneously predict semantic segmentations, motion field, and intermediate feature maps. The intermediate feature maps $z^{\mathrm{I}} \in \mathbb{R}^{512}$ are extracted based on the output of the middle convolutional layers before skip connections, and go through average pooling as well as linear layers to reduce feature channels with size of 512.

\section{Cross-modal Feature Fusion for 2D and 3D Microsurgical Anastomosis Trajectory Anticipation}
With diverse visual representations derived from microsurgical videos, future trajectory anticipation can be performed based on the fusion of multi-modal feature maps, which provides opportunities to assist operations under safe control and improve surgical training for automatic action generation. In this section, we introduce the proposed cross-modality feature fusion network to predict fused feature maps in a progressive coarse-to-fine manner with a designed long-short-term memory mechanism for the capability of temporal dependency. Based on the fine feature fusion, the network anticipates 2D and 3D trajectories to assist surgeons in intraoperative decisions and guide the robot to generate accurate actions.

\begin{figure*}[!htbp]
    \centering
    \includegraphics[width=1.0\linewidth]{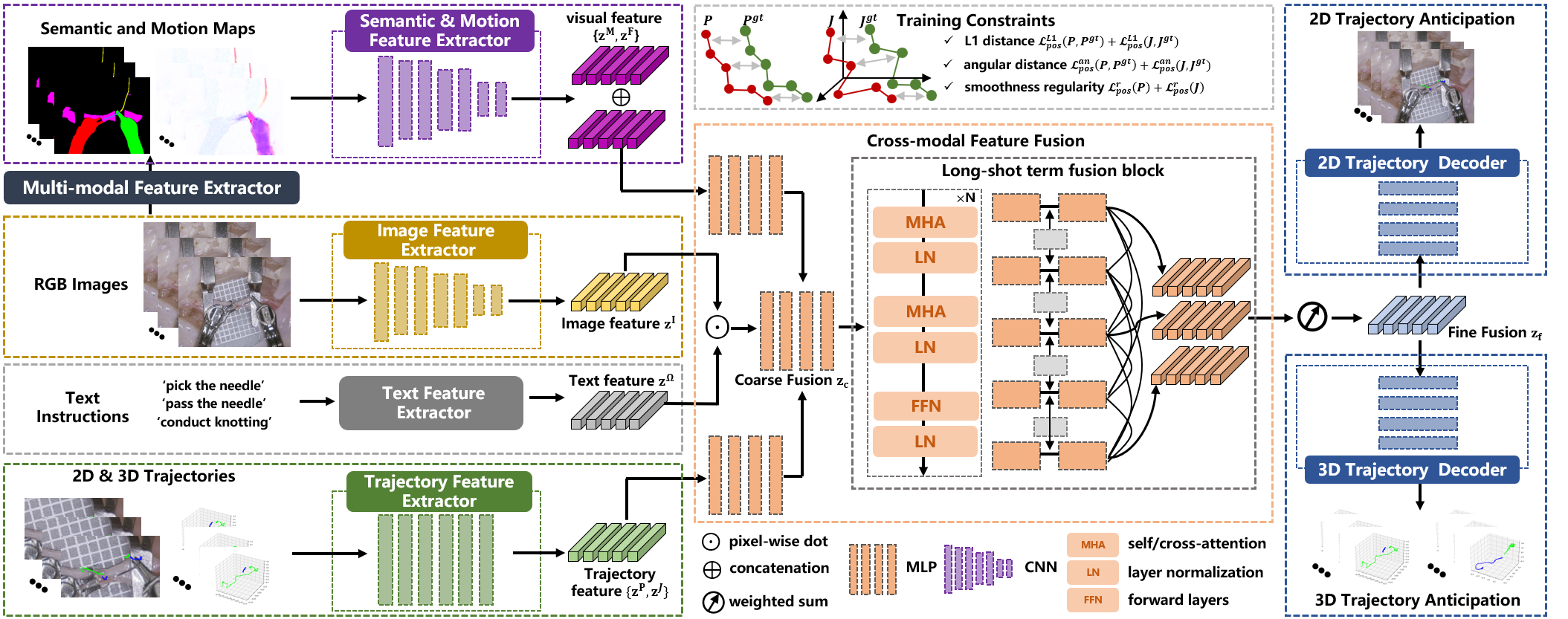}
    \caption{Illustration of the proposed cross-modal representation fusion network. The network takes the image feature, text feature, observed trajectory feature, and the combination of semantic and motion guidance as the input, and implements a coarse-to-fine cross-modal feature fusion scheme to generate the fused representation. After that, a 2D trajectory decoder and a 3D trajectory decoder is utilized to anticipate surgical trajectories in the future. During the training stage, a long-short-term fusion block is introduced to progressively capture temporal dependencies, and three constraints, including a L1 distance term, an angular distance term, and a smoothness regularity term,  are imposed on the predicted trajectories to improve the accuracy.}
    \label{fig:fusion_net}
    \vspace{-10pt}
\end{figure*}

\begin{figure}[!htbp]
    \centering
    \includegraphics[width=1.0\linewidth]{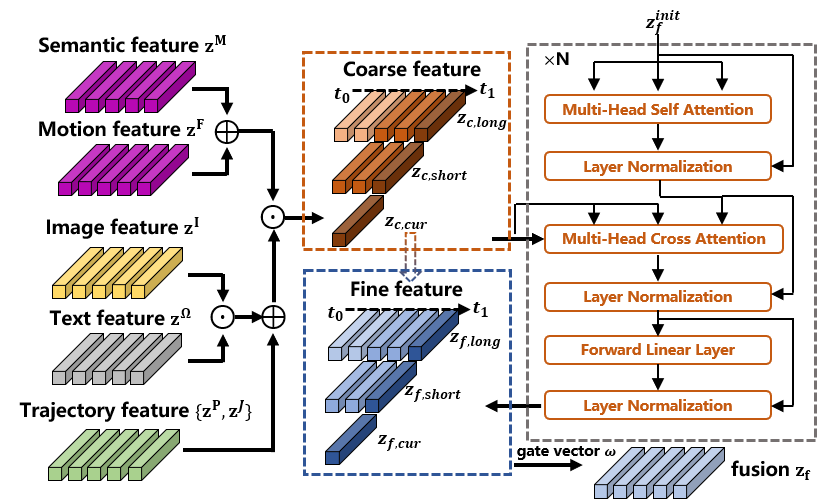}
    \caption{Detailed architecture of the progressive fusion block. Given the extracted semantic feature, motion feature, image feature, text feature, and trajectory feature, the fusion block takes the calculated coarse fused feature maps as input and generates the fine feature maps with respect to different periods in a progressive manner. After that, a learnable gate vector is implemented to alleviate weights for the final output.}
    \label{fig:fusion_scheme}
    \vspace{-10pt}
\end{figure}

\subsection{Problem Definition}
As shown in Fig. \ref{fig:visual_feature}, we have extracted semantic and motion representations with our proposed feature extractor $\mathcal{F}_\theta(\cdot)$. In addition, $\mathrm{\Omega}$ represents the current surgical phase in terms of text instructions. The combination of auxiliary representations is defined as $\mathrm{\hat{Z}}=\mathrm{Z}\bigcup\{\Omega\}=\{\mathrm{M}^{l}, \mathrm{M}^r, \mathrm{M}^{nd}, \mathrm{M}^{tr}, \mathrm{M}^{vl}, \mathrm{F},\mathrm{\Omega}\}$. For the 2D and 3D surgical trajectories, $\mathrm{P} \in \mathbb{R}^{2}$ represents the 2D pose sequences of left instrument $\mathrm{P}^l$  and right instrument $\mathrm{P}^r$ on the image plane, $\mathrm{J} \in \mathbb{R}^{3}$ denotes the 3D pose sequences of left instrument $\mathrm{J}^l$ and right instrument $\mathrm{J}^r$ under the robot coordinate system. We formulate $\mathrm{Y}=\{\mathrm{P}^l,\mathrm{P}^r,\mathrm{J}^l,\mathrm{J}^r\}$ for elements of future 2D and 3D trajectory anticipation with the cross-modality fusion network $\mathcal{H}_\theta(\cdot)$.  Our proposed network takes surgical video $\mathrm{I}$, observed surgical trajectories $\mathrm{Y}_{ob}$, and extracted auxiliary representations $\mathrm{\hat{Z}}$ observed from timestep $t_0$ to $t_1$ as input, and anticipates future trajectories $\mathrm{Y}$ from timestep $t_1+1$ to $t_2$. The training procedure of the multi-modal fusion network can be defined as:
\begin{equation}
    \mathcal{H}_{{\theta}^*}=\arg \min_{\theta} \mathcal{L}_{fuse}(\mathrm{I},\mathrm{\hat{Z}},\mathrm{Y}_{ob},\mathrm{Y},\mathrm{Y}^{gt}),
\end{equation}
where $\mathcal{L}_{fuse}$ represents the training constraint for multi-modal feature fusion and $\mathrm{Y}^{gt}$ refers to the ground truth trajectories demonstrated by human.

\subsection{Network Architecture}
As shown in Fig. \ref{fig:fusion_net}, the network for cross-modal feature fusion consists of two parts, i.e., the fusion module and the regression head, and the entire process of the cross-modality fusion is designed to be in a coarse-to-fine manner. A pretrained CLIP model  \cite{radford2021clip}  is implemented to extract text feature maps based on phase-specific instructions. Two convolutional layers are implemented to extract high-level feature maps of semantic segmentations and motion fields, and several linear layers are introduced to unify the feature dimension to be the same. The image feature maps are directly obtained from the pretrained feature representation networks mentioned above. The trajectory feature maps are composed by several linear layers taking the observed 2D and 3D trajectories as input. After conducting pixel-wise attention on all feature maps, the coarse fusion is obtained with a size of 512. After that, a transformer-based network is implemented to refine the fused feature maps, which is composed of the self-attention layer, multi-head attention layer, layer normalization, and the linear layer. To leverage the inherent relationships of temporal dependencies, a shared feature decoder is implemented to generate hierarchical refined feature maps for both the short and the long periods, and a learnable gate is designed to generate the fine fusion. Two separate regression decoders are implemented based on linear layers to predict the 2D and 3D surgical trajectories.

\subsection{Training Constraints}
Given the frame-wise multi-modal representations extracted by the proposed feature extractor $\mathcal{F}(\cdot)$, our proposed fusion framework performs temporal fusion and anticipates the future instrument trajectories. The first step is to efficiently fuse diverse representations. The semantic segmentations  $\{\mathrm{M}^l, \mathrm{M}^r, \mathrm{M}^{nd}, \mathrm{M}^{tr}, \mathrm{M}^{vl}\} \in \mathbb{R}^{H\times W}$ are concatenated together to form a 3D vector
$
\mathrm{M}^{c}=\mathrm{M}^l \oplus \mathrm{M}^r  \oplus \mathrm{M}^{nd} \oplus \mathrm{M}^{tr} \oplus \mathrm{M}^{vl} .
$
The feature maps of the concatenated semantic segmentations and motion fields are extracted and denoted as $z^\mathrm{M} \in \mathbb{R}^{512}$ and $z^\mathrm{F} \in \mathbb{R}^{512}$. Given the pre-defined surgical phase description $\mathrm{\Omega}$, we extract language feature maps noted as $z^\mathrm{\Omega}$. 
The 2D and 3D pose encoders extract positional features of two instruments represented as $z^\mathrm{P} \in \mathbb{R}^{512}$ and $z^{\mathrm{J}} \in \mathbb{R}^{512}$. We design the desired two coarse fused feature representations denoted as $\{z_{c}^{\mathrm{2D}}, z_{c}^{\mathrm{3D}}\}$, which can be derived by conducting cross-attention on feature vectors $\{z^{\mathrm{I}},z^{\mathrm{M}},z^{\mathrm{F}},z^{\mathrm{\Omega}},z^{\mathrm{P}},z^{\mathrm{J}}\}$:
\begin{equation}
    \begin{cases}
        z_{c}^{\mathrm{2D}} = ((z^{\mathrm{I}} \odot z^{\mathrm{\Omega}})\oplus z^{\mathrm{P}}) \odot (z^{\mathrm{M}} \oplus z^{\mathrm{F}})  \in \mathbb{R}^{512} \\
        z_{c}^{\mathrm{3D}} = ((z^{\mathrm{I}} \odot z^{\mathrm{\Omega}})\oplus z^{\mathrm{J}}) \odot (z^{\mathrm{M}} \oplus z^{\mathrm{F}})  \in \mathbb{R}^{512} \\
    \end{cases}.
    \label{feature_concat}
\end{equation}
It should be noted that the coarse feature maps are based on frame level, and no temporal fusion is introduced in this procedure. Consequently, to improve the temporal capability, we implement a transformer-based network to conduct temporal refinement of the coarse feature representations. Given coarse feature maps $z_c=\{z_c^{2D}, z_c^{3D}\}$ from timestep t$_0$ to t$_1$, and the refined feature maps $z_f=\{z_f^{2D}, z_f^{3D}\}$ can be represented as:
\begin{equation}
    \begin{cases}
        z_{f}^{\mathrm{2D}} = \mathrm{Trans}(z_{c,t_0}^{\mathrm{2D}},z_{c,t_0+1}^{\mathrm{2D}},...,z_{c,t_1}^{\mathrm{2D}}) \\
        z_{f}^{\mathrm{3D}} = \mathrm{Trans}(z_{c,t_0}^{\mathrm{3D}},z_{c,t_0+1}^{\mathrm{3D}},...,z_{c,t_1}^{\mathrm{3D}})
    \end{cases}.
\end{equation}
In details, given the initialized future feature sequence $z_f^{init}=\{z_{f,t_1+1}^{init},z_{f,t_1+2}^{init}...,z_{f,t_2}^{init}\}$ , the refine process of the transformer network $\mathrm{Trans(\cdot)}$ can be represented as:
\begin{equation}
    \begin{cases}
       z_f^{1} = \mathrm{LN}(z_f^{init}+\mathrm{MHA}(z_f^{init}+\mathrm{Pos}(z_f^{init}),z_f^{init}))\\
       z_f^2=\mathrm{LN}(z_f^1 + \mathrm{MHA}(z_f^1+\mathrm{Pos}(z_f^{init}), z_c+\mathrm{Pos}(z_c),z_c))\\
       z_f = \mathrm{LN}(z_f^2 + \mathrm{FFN}(z_f^2)) 
    \end{cases},
\end{equation}
where $\mathrm{MHA}(\cdot)$ refers to the self-attention and cross-attention mechanisms, $\mathrm{LN}(\cdot)$ refers to the layer normalization, and $\mathrm{FFN}(\cdot)$ refers to the forward linear layers with the positional encoding function $\mathrm{Pos}(\cdot)$.
To improve the effectiveness of temporal feature aggregation for diverse modalities, we introduce a long-short-term fusion mechanism that can jointly emphasize short-term observation and leverage long-range historical context. Conventional temporal fusion approaches leverage equal attention for a certain temporal sliding window, which hinders the capture of local motion dynamics and long-term dependencies. To address the limitation, our proposed fusion scheme introduces a learnable gate on aggregating short-term feature maps and alleviating the long-term feature maps as auxiliary. As shown in Fig. \ref{fig:fusion_scheme}, three different feature sets $\{z_{f,cur},z_{f,short},z_{f,long}\}$ with different time periods ranging from the current observation to the complete temporal sequence are generated, and a learnable gate vector $\omega$ is introduced to alleviate weights for different terms:
\begin{equation}
    \begin{cases}
       z_{f,cur} = \mathrm{Trans}(z_{c,t_1})\\
       z_{f,short} = \mathrm{Trans}(z_{c,t_1-s},...,z_{c,t_1})\\
       z_{f,long} = \mathrm{Trans}(z_{c,t_0},...,z_{c,t_1})\\
       \alpha_z = \mathrm{Softmax}(\omega(z_{f,cur}\oplus z_{f,short}\oplus z_{f,long}))\\
       z_f = \alpha_z (z_{f,cur}\oplus z_{f,short}\oplus z_{f,long})
    \end{cases}.
\end{equation}

After that, the proposed 2D and 3D trajectory decoders $\{\mathrm{Dec^{2D}(\cdot)},\mathrm{Dec^{3D}(\cdot)}\}$ take the refined feature maps $\{z_f^{\mathrm{2D}},z_f^{\mathrm{3D}}\}$ as input and predict the future trajectories from timestep $t_1+1$ to $t_2$:
\begin{equation}
    \begin{cases}
        \{\mathrm{P}_{t_1+1},\mathrm{P}_{t_1+2},...,\mathrm{P}_{t_2}\} = \mathrm{Dec^{2D}}(z_f^{\mathrm{2D}}) \\
        \{\mathrm{J}_{t_1+1},\mathrm{J}_{t_1+2},...,\mathrm{J}_{t_2}\} = \mathrm{Dec^{3D}}(z_f^{\mathrm{3D}}) 
    \end{cases}.
\end{equation}
The training constraint is defined to calculate the distance between predicted future and ground truth human-demonstrated trajectories noted as $\mathcal{L}_{pos}$. Inspired by \cite{ma2025madiff}, the trajectory similarity term is represented by the weighted sum of L1 distance $\mathcal{L}_{pos}^{L1}$, angular similarity $\mathcal{L}_{pos}^{an}$, and smoothness regularity $\mathcal{L}_{pos}^{r}$. The L1 distance term calculates the Euclidean distance between the predicted 2D and 3D trajectories and the ground truth, and the angular similarity term implements the cosine similarity function $\mathrm{CosSim(\cdot)}$, forcing the network to generate trajectories with the correct directions instead of only low distance errors, and the smoothness term promises the stability and continuity of the prediction:
\begin{equation}
    \begin{cases}
        \mathcal{L}_{pos}^{L1} = \frac{ \sum_{i=(t_1+1)}^{t_2} (||\mathrm{P}_i - \mathrm{P}_i^{gt}||_2 + \alpha_{pos}^{3D}||\mathrm{J}_i - \mathrm{J}_i^{gt}||_2)} {t_2-(t_1+1)}\\ 
        \mathcal{L}_{pos}^{an} = \frac{  \sum_{i=t_1}^{t_2} \mathrm{CosSim}(\mathrm{P}_i-\mathrm{P}_{i+1},\mathrm{P}_i^{gt}-\mathrm{P}_{i+1}^{gt})} {t_2-t_1} + \\ ~~~~~~~~~~  \alpha_{pos}^{3D}\frac{  \sum_{i=t_1}^{t_2}  \mathrm{CosSim}(\mathrm{J}_i-J_{i+1},\mathrm{J}_i^{gt}-\mathrm{J}_{i+1}^{gt}) } {t_2-t_1} \\
        \mathcal{L}_{pos}^r =  \frac{  \sum_{i=t_1}^{t_2} (||\mathrm{P}_i-\mathrm{P}_{i+1}||_2 + \alpha_{pos}^{3D}||\mathrm{J}_i-\mathrm{J}_{i+1}||_2 )} {t_2-t_1}
    \end{cases}.
\end{equation}
The combined training loss function for multi-modal feature fusion can be defined as:
\begin{equation}
        \mathcal{H}_{\theta^*} = \arg \min_{\theta} \mathcal{L}_{pos}   = \alpha_{pos}^{L1}\mathcal{L}_{pos}^{L1} + \alpha_{pos}^{an}\mathcal{L}_{pos}^{an} + \alpha_{pos}^{r}\mathcal{L}_{pos}^{r}.
\end{equation}

In summary, we follow a two-stage training strategy for our proposed framework. We train the multi-modal feature extraction network $\mathcal{F}(\cdot)$ following the training constraint $\mathcal{L}_{feat}$. After that, we freeze the parameters of it and use the pretrained model to extract feature representations. Our proposed cross-modal fusion network $\mathcal{H}(\cdot)$ is trained based on the training constraint $\mathcal{L}_{pos}$.

\section{Experimental Settings}

\subsection{Dataset Description}
\subsubsection{Phantom Dataset (D-I)}
The phantom dataset originally contains 100 video sequences for suture making and 60 video sequences for knotting with about 76k frames in total. For pretraining the feature extractor, 4.5k frames are annotated of instrument masks with 2.7k frames for training and 1.8k frames for evaluation, 3.6k frames are annotated of needle masks with 3k for training and 0.6k for evaluation, 1.9k frames are annotated of thread masks with 1.5k for training and 0.4k for evaluation, and 1.4k frames are annotated of target vessel masks with 1k for training and 0.4k for evaluation. In addition, 1.2k frames with manually annotated 2D pose are provided for training the detection network. For trajectory anticipation, after preprocessing the extracted 2D and 3D trajectories, there are 56k samples for training and 4k samples for validation.

\subsubsection{Ex-vivo Dataset (D-II)}
The ex-vivo dataset originally contains 30 video sequences both for suture making and knotting with 51k frames in total. For pretraining the feature extractor, 3.4k frames are annotated of instrument masks with 2k frames for training and 1.4k frames for evaluation, 3.3k frames are annotated of needle masks with 2.5k for training and 0.8k for evaluation, 2.1k frames are annotated of thread masks with 1.6k for training and 0.5k for evaluation, and 2.4k frames are annotated of target vessel masks with 2k for training and 0.4k for evaluation. In addition, 0.7k frames with manually annotated 2D pose are provided for training the detection network. For trajectory anticipation, after preprocessing the extracted 2D and 3D trajectories, there are 40k samples for training and 2k samples for validation.
\subsection{Implementation Details}

We first pretrain the multi-modal feature extraction network on the tasks of semantic segmentation and motion prediction following the training constraint $\mathcal{L}_{feat}$. After that, the parameters of the feature extractor are frozen and the derived feature maps are fed into the cross-modal fusion network to regress the instrument trajectories following the constraint $\mathcal{L}_{pos}$. The first stage of multi-modal feature extraction is trained with 20 epochs, and the second stage of feature fusion and trajectory anticipation is trained with 10 epochs. 
Without further clarification, we sample the image sequence at 1 Hz with the temporal perception field of 5s. The 2D and 3D trajectories are sampled at 10 Hz and we anticipate the future state for 5s.
The proposed framework is implemented by PyTorch and is trained on three NVIDIA TITAN RTX 3090 GPUs with 24 GB memory. The coefficients in the training constraint $\mathcal{L}_{feat}$ are set as $\alpha_{seg}^{ins}=1.0, \alpha_{seg}^{nd}=5.0, \alpha_{seg}^{tr}=5.0,\alpha_{seg}^{vl}=1.0, \alpha_{mot}^s=1.0, \alpha_{mot}^r=10.0$, and in $\mathcal{L}_{pos}$ are set as $\alpha_{pos}^{L1}=0.2, \alpha_{pos}^{an}=0.01, \alpha_{pos}^r=0.05,  \alpha_{pos}^{3D}=5.0$. All coefficients are selected based on parameter ablations and are chosen for the best performance. All images are resized to $320\times256$ and normalized from 0 to 1 by dividing the maximum value.

\begin{table*}[!htbp]
\centering
\scriptsize
\renewcommand\arraystretch{1.5}
\caption{ Comparative results of semantic segmentation with existing methods}
\begin{tabular}{lccccccccc}
\hline
 \multirow{2}{*}{Approaches} & \multirow{2}{*}{Datasets}& \multicolumn{2}{c}{Instruments} & \multicolumn{2}{c}{Needles}  & \multicolumn{2}{c}{Threads} & \multicolumn{2}{c}{Vessels} \\
& & IOU (\%) & Dice (\%) & IOU (\%) & Dice (\%) & IOU (\%) & Dice (\%) & IOU (\%) & Dice (\%) \\
  \hline
  Segformer \cite{xie2021segformer} & \multirow{5}{*}{D-I} &   95.44 & 97.64   & 59.72   & 74.13   & 52.00   &  62.01   &  94.94  & 97.40  \\
  TPSIS \cite{zhou2023text} & & 92.13   & 95.83   &  55.74  & 70.66   & 47.77   & 58.54    & 94.17   &96.98    \\
TAPIS \cite{ayobi2025pixel} & &  95.15  & 97.48   &  57.70  &  72.34  &  \underline{55.47}  & \underline{64.75}    & 96.09   & 98.00   \\
 SurgicalSAM \cite{yue2024surgicalsam} &  & \underline{96.60}   & \underline{98.24}   &  \underline{61.13}  &  \underline{75.21}  & 50.32   & 60.33    & \underline{97.96}   &  \underline{98.97}    \\
 Ours &  & \textbf{97.58}    & \textbf{98.76}    & \textbf{67.70}   & \textbf{80.25}   & \textbf{57.91}   & \textbf{66.32}    & \textbf{98.21}   & \textbf{99.09}    \\
\hline
 Segformer \cite{xie2021segformer} & \multirow{5}{*}{D-II} & 93.73   &  96.69  & 75.88   & 85.72   &  55.56  & 70.01    & 61.26    & 74.32  \\
  TPSIS \cite{zhou2023text} & &  86.68  & 92.69   &  75.71  & 85.58   &  49.29  & 64.51    & 53.16   & 67.39   \\
TAPIS \cite{ayobi2025pixel} & &  93.72  &  96.69  & \underline{77.34}   &  \underline{86.62}  &  \underline{59.66}  &  \underline{73.33}   &  61.46  & 75.69   \\
 SurgicalSAM \cite{yue2024surgicalsam} &  & \underline{96.06}   &  \underline{97.96}  & 76.95   &  86.20  &  56.58  & 70.44    &  \underline{62.10}  & \underline{75.70}     \\
Ours &  & \textbf{96.49}  & \textbf{98.17}   & \textbf{82.08}   & \textbf{89.55}   & \textbf{66.01}   & \textbf{78.06}     & \textbf{66.98}    &  \textbf{79.25}       \\
\hline
 Segformer \cite{xie2021segformer}  & \multirow{5}{*}{Average} &  94.70  & 97.23   & 68.39   &  80.35  & 54.03   &  66.58   & 68.21   & 79.09  \\
  TPSIS \cite{zhou2023text} & &  89.78  & 94.48   & 66.45   & 78.66   & 48.64   & 61.95     &  61.63 &  73.50  \\
TAPIS \cite{ayobi2025pixel} & &  94.53  &  97.14  &  68.24  &  80.00  & \underline{57.86}   &\underline{69.65}     & 68.61   & 80.30   \\
 SurgicalSAM \cite{yue2024surgicalsam} &  & \underline{96.37}   & \underline{98.12}   & \underline{69.62}   &   \underline{81.11} & 53.90   &  66.10   &  \underline{69.51}  &  \underline{80.51}    \\
Ours &  &  \textbf{97.11}  &  \textbf{98.51}  &  \textbf{75.41}  &  \textbf{85.24}  & \textbf{62.54}   & \textbf{73.02}    & \textbf{73.43}   & \textbf{83.35}  \\
\hline

\label{tab:quan_MMFeatNet_seg}
\end{tabular}
\end{table*}

\begin{table*}[!htbp]
\centering
\scriptsize
\renewcommand\arraystretch{1.5}
\caption{ Comparative results of motion estimation with existing methods}
\begin{tabular}{lcccccc}
\hline
\multirow{2}{*}{Approaches}& \multicolumn{2}{c}{D-I} & \multicolumn{2}{c}{D-II} & \multicolumn{2}{c}{Average}   \\
  & PSNR (dB) & SSIM (\%) & PSNR (dB) & SSIM (\%) & PSNR (dB) & SSIM (\%)  \\
  \hline
EndFlow \cite{yang2021dense}   & 32.48   & 95.34   & 31.95 & 93.17  & 32.38 & 94.52    \\
RAFT \cite{teed2020raft}  & \underline{33.72}   & \underline{96.17}   & 32.33  & 93.89  & 33.20 &  95.31   \\
GMA \cite{jiang2021learning} & 33.34  &  96.07  & \underline{33.39} &  \underline{94.89} & \underline{33.36} &  \underline{95.63}   \\
Ours  & \textbf{35.62}  & \textbf{97.09}   & \textbf{34.10}  & \textbf{96.08}   & \textbf{35.05}    & \textbf{96.71}      \\
\hline
 
\label{tab:quan_MMFeatNet_mot_phase}
\end{tabular}
\end{table*}

\subsection{Evaluation Metrics}
For the task of semantic segmentation including surgical instruments, needle, thread, and operated vessel, we report Intersection Over Union (IOU$\uparrow$) and Dice coefficient (Dice$\uparrow$). These two metrics are both calculated based on the similarity between predicted binary masks and annotations. For the task of motion estimation, we report Peak Signal-to-Noise Ratio (PSNR$\uparrow$) and Structure Similarity Index Measure (SSIM$\uparrow$) to evaluate the difference between target images and transformed images by the predicted motion field. For the task of trajectory anticipation, we report Average Displacement Error (ADE$\downarrow$) and Final Displacement Error (FDE$\downarrow$) to calculate the distance between predicted trajectories and ground truths. In addition, to evaluate the geometrical similarity between two temporal trajectories, we introduce the Frechét Distance (FRD$\downarrow$) and the Hausdorff Distance (HD$\downarrow$). Our reported results for 2D trajectory are based on the unit of pixel and for 3D trajectory are based on the unit of  $10^{-2}$ mm.

\subsection{Comparison Methods}
To demonstrate the superior performance of our proposed framework against existing methods, we have conducted diverse comparison experiments on different tasks. On the one hand, we compare our proposed framework for multi-task feature extraction on the task of semantic segmentation with Segformer \cite{xie2021segformer}, TPSIS \cite{zhou2023text}, TAPIS \cite{ayobi2025pixel}, and SurgicalSAM \cite{yue2024surgicalsam}, and compare with EndFlow \cite{yang2021dense}, RAFT \cite{teed2020raft}, and GMA \cite{jiang2021learning} on the task of motion estimation. On the other hand, we compare our framework for surgical trajectory anticipation with diverse existing methods including BC \cite{florence2022ibc}, TTM \cite{shi2022ttm}, LED \cite{mao2023led}, USST \cite{bao2023usst}, SDMamba \cite{wang2025sdmamba}, iDPOE \cite{wang2025idpoe}, and MADiff \cite{ma2025madiff}. Our comparative methods cover an extensive range from time series forecasting to surgical and natural trajectory anticipation. All comparison methods are trained based on our dataset following the implementation described in their works, and both quantitative and qualitative results are reported.

\subsection{Ablation Study}
To prove the effectiveness of each blocks in our proposed framework and training strategies, we have conducted diverse ablated experiments from several different aspects. For the network structure, we evaluate the influence of different components of visual representations. For the fusion scheme, we evaluate the influence of different cross-modality fusion mechanisms and conduct diverse parameter ablations for the network training. We also evaluate the influence of the length of the anticipation period and report action-specific results to discuss the complexity of different operations.

\begin{figure*}[!htbp]
    \centering
    \includegraphics[width=1.0\linewidth]{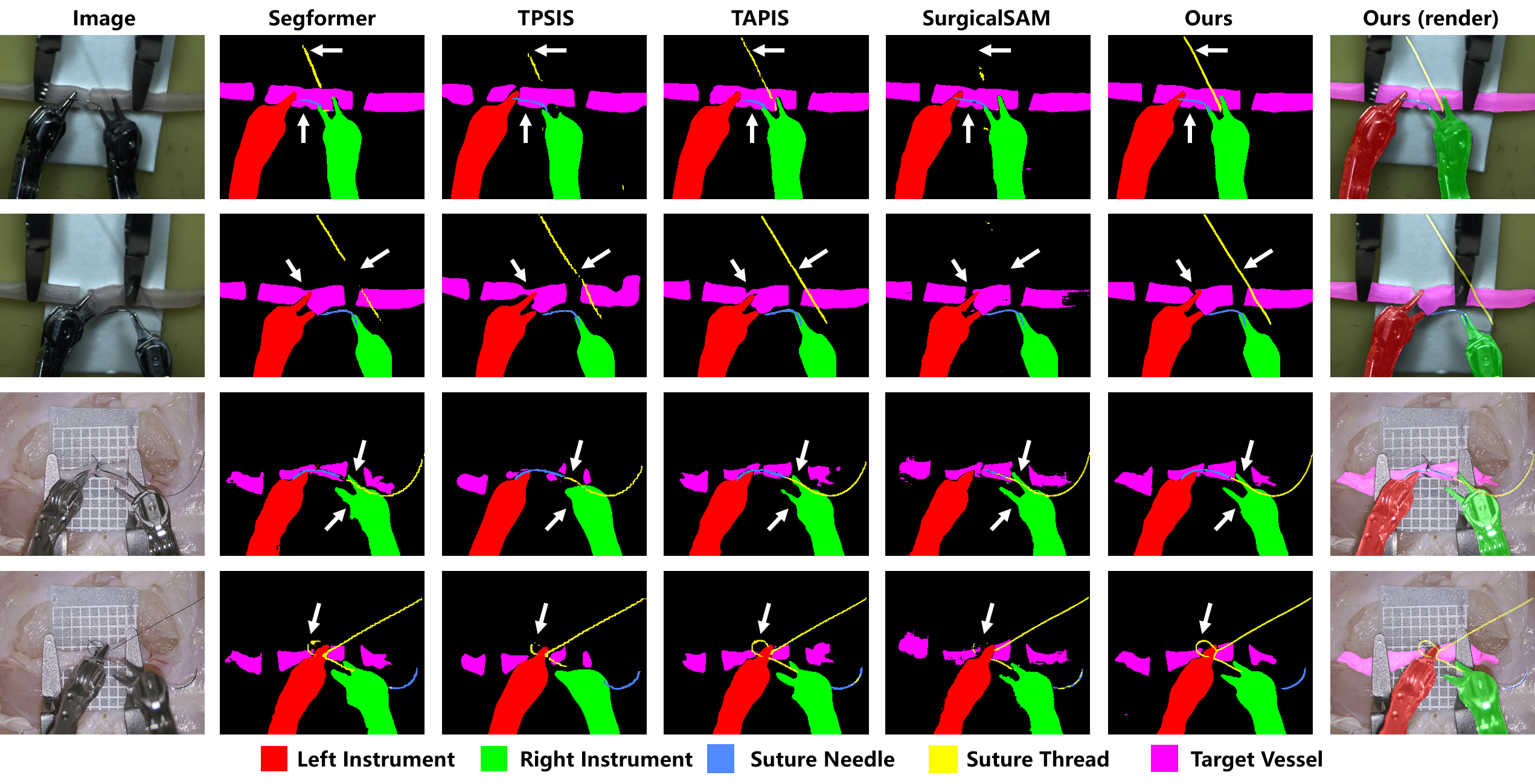}
    \caption{Qualitative results of semantic segmentation with different methods on both phantom and ex-vivo datasets. The five classes are visualized with different colors, and the white arrows indicate the regions with significant improvements by our method.}
    \label{fig:result_multi_seg}
\end{figure*}

\section{Comparison Results and Analysis}
Without further clarification, the \textbf{bold} and \underline{underline} results in tables refer to the best and second best performance for each experiment, respectively. In this section,  we first compare our proposed multi-feature extraction network with existing methods on semantic segmentation and motion estimation.  And then we compare our proposed cross-modal fusion framework with existing approaches for 2D and 3D trajectory anticipation based on extracted semantics and motion maps.

\subsection{Multi-modal Feature Representation Extraction on Semantic Segmentation and Motion Estimation}
The comparative results of semantic segmentation are reported in Table \ref{tab:quan_MMFeatNet_seg}, and two metrics IOU and Dice are evaluated for the recognition of the surgical instrument, suturing needle, thread, and target vessel. It can be found that our proposed network achieves the best performance on all metrics, which demonstrates the effectiveness of the multi-task learning scheme and the inherent relationships among different surgical objects contribute to the superior performance compared with the individual learning. The first part reports the results on the phantom dataset D-I, and our proposed method achieves an improvement of 1.01\% and 0.53\% on IOU and Dice for instrument recognition, respectively, and 10.75\% and 6.70\% for the needle, and 4.40\% and 2.42\% for the thread, and 0.26\% and 0.12\% for the vessel compared to the second-best approach. The second part refers to the results on the ex-vivo dataset D-II. Our method achieves an improvement of 0.45\% and 0.21\%  for the instrument, and 6.13\% and 3.38\% for the needle, and 10.64\% and 6.45\% for the thread, and 7.86\% and 4.69\% for the vessel. The third part reports the average performance on those two datasets, and it can also be found that our proposed method achieves the best performance on all objects. Several qualitative examples are provided in Fig. \ref{fig:result_multi_seg}, where five different colors represent five classes. The first column shows the input image, the second to fifth columns visualize four comparative results, and the last two columns refer to our prediction as well as semantics overlaid on the original images. The white arrows indicate the significant improvements compared with other methods. It can be found that the recognition of detailed objects, especially for the needle and thread, is obviously improved via introducing the semi-supervised multi-task learning scheme, and the structure of large objects, including instruments and vessels, is also perfectly recognized.

In Table \ref{tab:quan_MMFeatNet_mot_phase}, we report two metrics PSNR and SSIM to conduct comparative experiments between our method and three existing approaches on the task of motion estimation. On the phantom dataset D-I, our proposed method achieves a performance improvement of 5.63\% and 0.96\% on PSNR and SSIM, respectively, and on the ex-vivo dataset D-II, our method achieves the improvement of 2.13\% and 1.25\% compared with the second-best method. On the average performance of those datasets, our proposed method achieves an improvement of 5.07\% and 1.13\% on PSNR and SSIM. From the reported results, it can be found that our proposed framework gains the best performance on all metrics with the most accurate motion estimation.

\begin{table*}[!htbp]
\centering
\scriptsize
\renewcommand\arraystretch{1.5}
\caption{ Comparative results of 2D and 3D surgical trajectory anticipation with existing methods}
\begin{tabular}{lccccccccc}
\hline
\multirow{2}{*}{Approaches}& \multirow{2}{*}{Datasets}& \multicolumn{4}{c}{2D Trajectory } & \multicolumn{4}{c}{3D Trajectory  } \\
  & & ADE & FDE & FRD & HD & ADE & FDE & FRD & HD\\
  \hline
  BC \cite{florence2022ibc} & \multirow{8}{*}{D-I} &   14.21 &   15.71 & 19.20  & 14.24  & 57.00  &  72.92 & 87.93  & 58.33    \\
  TTM \cite{shi2022ttm}  &  & \underline{12.32}   &15.69    & \underline{18.44}  & 13.13  & 29.13  & 47.55  & 55.03  & 38.38    \\
LED \cite{mao2023led}  &  & 13.83   & 17.07   & 24.19  & 20.82  & 26.88  & 44.86  & 52.25  &  40.48   \\
USST \cite{bao2023usst}    & & 14.80   &  17.37  & 20.36  & 15.21  &  33.85 & 55.85  & 66.95  & 37.98    \\
SDMamba \cite{wang2025sdmamba} &  & 13.50   & \underline{15.59}   & 18.88  & \underline{12.01}  &  42.58 &  58.62 & 73.96  & 41.46     \\
iDPOE \cite{wang2025idpoe}  &  & 13.40   & 17.48   &  20.44 &  15.58 & 27.65  & 45.93  & \underline{51.47}  &  38.30   \\
MADiff \cite{ma2025madiff}  & & 13.13   & 17.43   & 19.54  & 12.42  & \underline{26.50}  & \underline{41.73}&  56.67 & \textbf{32.15}    \\
Ours  & &  \textbf{11.20}  & \textbf{13.81}   & \textbf{15.63}  & \textbf{11.28} & \textbf{25.28}   & \textbf{40.83}  & \textbf{46.74} & \underline{34.13}      \\
\hline
 BC \cite{florence2022ibc} & \multirow{8}{*}{D-II} &  22.52  & 23.07   &  31.33 &  17.65 & 50.82  & 65.20  & 93.28  & 33.80    \\
 TTM \cite{shi2022ttm}   &  & 20.75   &  25.41  &  30.14 &  15.27 & 40.61  & 58.59  & 79.28  &  38.84   \\
LED \cite{mao2023led}  &   &  18.63  &  24.38  &  37.59 &  29.05 & 41.22  &  61.36 & 77.51  & 39.18   \\
USST \cite{bao2023usst}&  &  16.14  & \underline{18.29}   & 24.27  & \underline{12.93}  & 41.73  & 61.85  & 78.74  & \textbf{34.35}  \\
SDMamba \cite{wang2025sdmamba} &  &  16.45  & 18.88   & 24.96  & 15.97  & 40.42  &  \underline{56.63} & \underline{74.49}  & 37.15    \\
iDPOE \cite{wang2025idpoe} &  &  16.87  & 20.01   & \underline{23.77}  & 13.95  & \underline{40.06}  &  62.21 & 75.21  & 36.42    \\
MADiff \cite{ma2025madiff} & & \textbf{14.14} & 19.83   & 27.22  & 16.84  &  42.31 & 63.96  & 79.86  & \underline{34.50}    \\
Ours  & & \underline{15.25}   & \textbf{16.29}  & \textbf{21.32} & \textbf{11.78} & \textbf{40.04}  & \textbf{55.12}  & \textbf{71.43} & 34.70     \\
\hline
 BC \cite{florence2022ibc} & \multirow{8}{*}{Average} & 17.15   & 18.31   &23.49   & 15.45  & 54.82  & 70.19  & 89.92  &  49.66   \\
 TTM  \cite{shi2022ttm}  &  & 15.30   & 19.13   & 22.58  & 13.89  & 33.19  & 51.45  & 63.60  &  38.54   \\
LED \cite{mao2023led}  &   & 15.53  & 19.65   & 28.93  & 23.73  & \underline{31.95}  & 50.69  & 61.18  & 40.02     \\
USST \cite{bao2023usst}&  & 15.27   & 17.70   & 21.74  & 14.40  & 36.64  & 57.97  &  71.12 & 36.70     \\
SDMamba \cite{wang2025sdmamba} &  &  14.54  & \underline{16.75}   & \underline{21.03}  &  \underline{13.41} & 41.82  &  57.92 &  74.15 & 39.94    \\
iDPOE \cite{wang2025idpoe} &  &  14.63 &  18.37  & 21.62  &  15.00 & 32.04  & 51.68  & \underline{59.86}  & 37.64   \\
MADiff \cite{ma2025madiff} & &  \underline{13.49}  & 18.28   & 22.25  &  13.98 &  32.09 & \underline{49.59}  & 64.87  &\textbf{32.98}      \\
Ours  & & \textbf{12.63}   & \textbf{14.69}   & \textbf{17.64}  & \textbf{11.46}  & \textbf{30.50}  & \textbf{45.88}  & \textbf{55.47}  & \underline{34.33}    \\
\hline
 
\label{tab:quan_MMFuseNet_traj}
\vspace{-10pt}
\end{tabular}
\end{table*}

\subsection{Cross-modal Feature Fusion on 2D and 3D Anastomosis Trajectory Anticipation}
In Table \ref{tab:quan_MMFuseNet_traj}, the comparative results on 2D and 3D trajectory anticipation are reported, and four metrics, including ADE, FDE, FRD, and HD, are implemented to provide a quantitative comparison. It should note that the metrics ADE and FDE focus on the point-wise distance error, and the metrics FRD and HD evaluate the similarity between two trajectories. It can be found that our proposed method achieves the best average performance for both 2D and 3D anticipation. The first part provides the comparative results on the phantom dataset D-I between our proposed method and five existing methods. Compared with the second-best performance, our proposed method gains a performance improvement of 9.10\%, 11.42\%, 15.24\%, and 6.08\% for the metrics ADE, FDE, FRD, and HD on the 2D case, respectively. The significant improvements on FRD and HD indicate that our prediction holds the most similar shape of future trajectories and preserves the topological structure. Meanwhile, our method also achieves an improvement of 4.60\%, 2.16\% and 9.19\% for ADE, FDE, and FRD on the 3D case. The second part reports the results on the ex-vivo dataset D-II, since the surgical environment in this dataset is much more complex than that in the phantom case, it is much more challenging for the framework to predict precise trajectories. It can be found that our method achieves a performance gain of 10.93\%, 10.31\%,  and 8.89\% for the metrics FDE, FRD, and HD on the 2D case, compared with the second-best approach, and 0.05\%, 2.67\%, and 4.11\% for the metrics ADE, FDE, and FRD on the 3D case. It can also be found that the performance of comparison methods varies a lot on the two datasets since the surgical scenarios and operations on the ex-vivo case are much more complicated than those on the phantom case, resulting in degraded performance with unstable anticipation adapted from the phantom case to ex-vivo case. We have also reported the average performance based on the two datasets, and the above quantitative results have proved that our proposed cross-modal fusion framework as well as the long-short term fusion mechanism contribute to the best performance compared with existing methods.

\begin{figure*}[!htbp]
    \centering
    \includegraphics[width=1.0\linewidth]{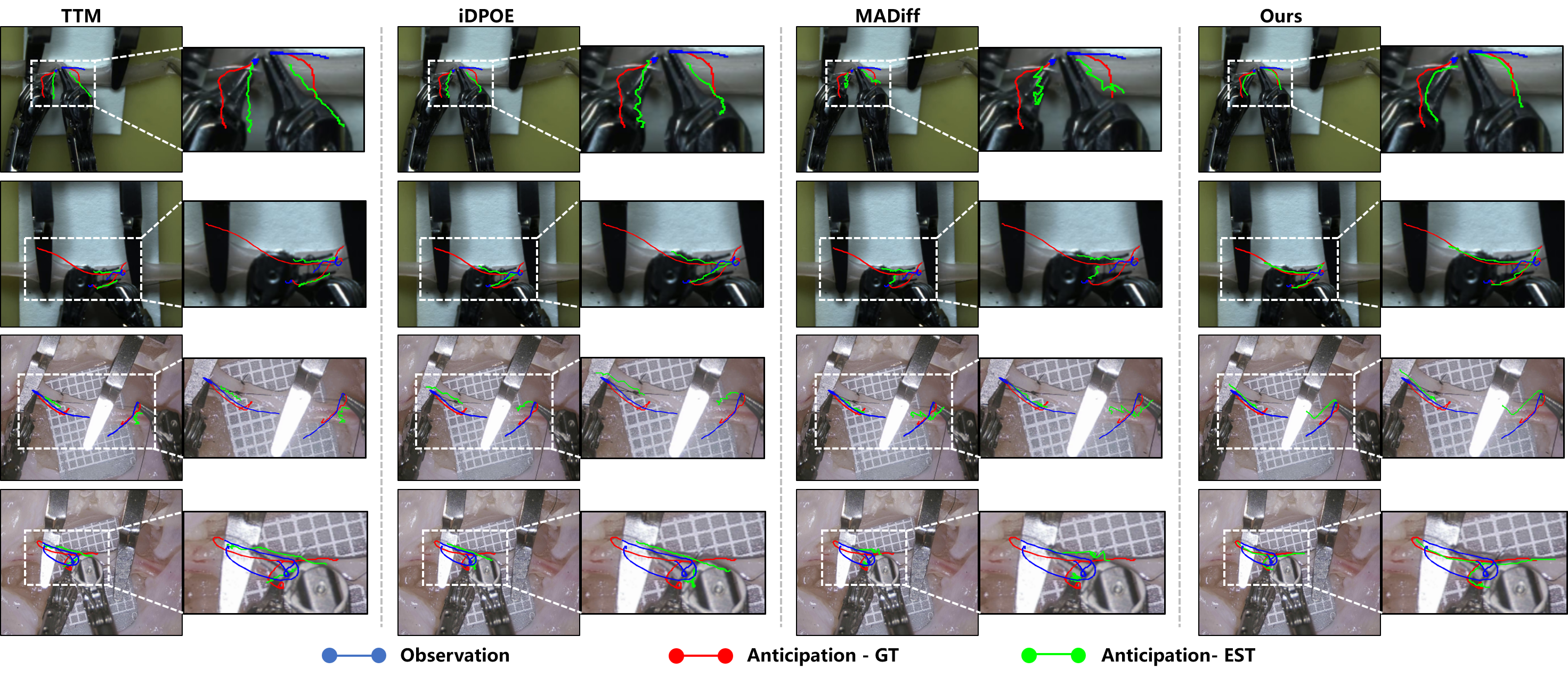}
    \caption{Qualitative results of 2D trajectory anticipation with different methods on both phantom and ex-vivo datasets. The blue line refers to the past observation, the red line refers to the ground truth collected by human demonstration, and the green line refers to the anticipation by different methods. The zoomed images visualized in the right part indicate the region of surgical instrument-tissue interactions.}
    \label{fig:result_2d_traj}
    \vspace{-10pt}
\end{figure*}

\begin{figure*}[!htbp]
    \centering
    \includegraphics[width=1.0\linewidth]{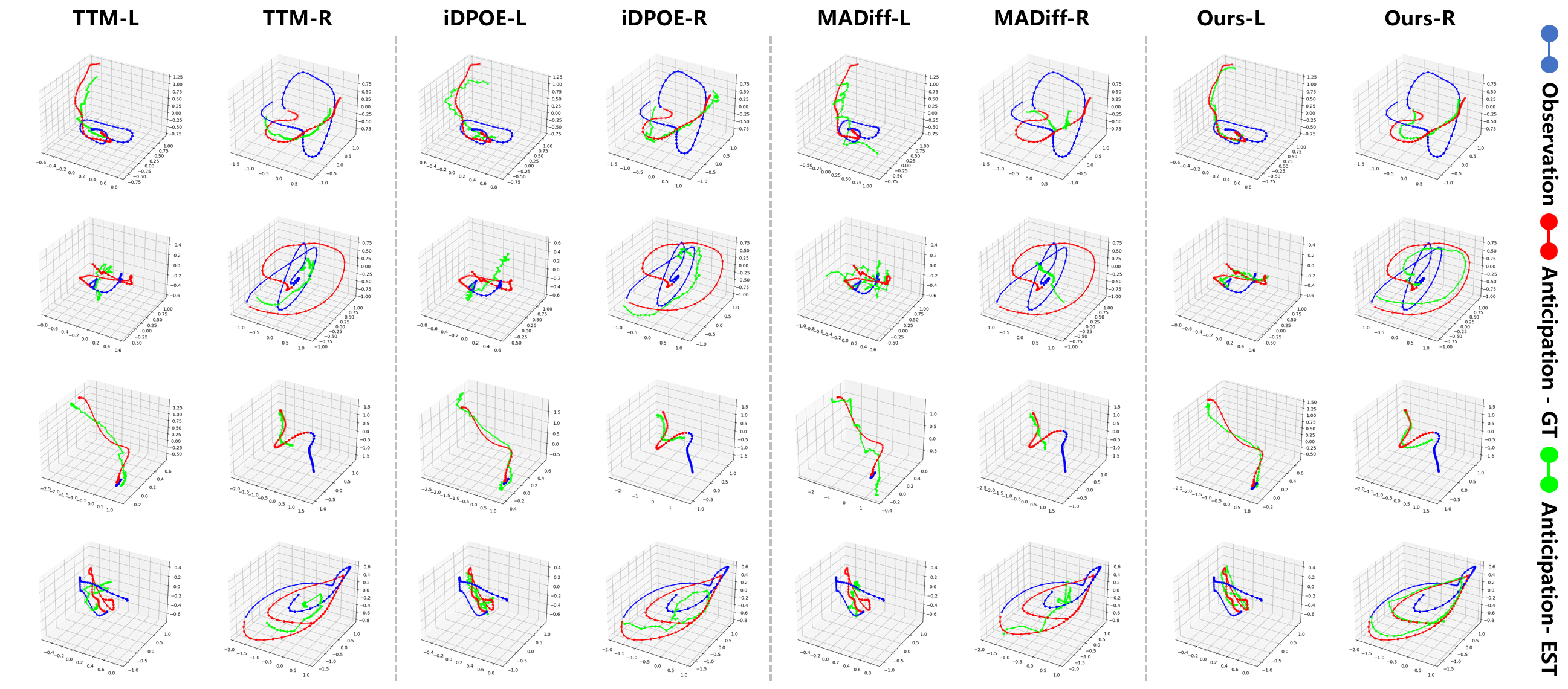}
    \caption{Qualitative results of 3D trajectory anticipation with different methods on both phantom and ex-vivo datasets. The blue line refers to the past observation, the red line refers to the ground truth collected by human demonstration, and the green line refers to the anticipation by different methods. The left and right instrument trajectories are visualized separately. }
    \label{fig:result_3d_traj}
    \vspace{-10pt}
\end{figure*}

The qualitative results on 2D trajectory anticipation are shown in Fig. \ref{fig:result_2d_traj}, where the blue color refers to the observed trajectories in the past 5s, the green color refers the prediction by different methods in the future 5s, and the red color refers to the ground truth trajectories demonstrated by human. Note that the visualized image is selected as the last frame of observation, and trajectories are overlaid on it with a zoomed-in part shown at the top left corner. It can be found that our proposed method anticipates the future trajectory with the best accuracy for both the final point position and the topological structure. The qualitative results on 3D trajectory prediction are visualized in Fig. \ref{fig:result_3d_traj}. For each method, the left column shows the 3D trajectory of the left instrument and the right column shows the right instrument. It can be found that our proposed method achieves the best performance with the least position error, and although the shape of the trajectory is slightly different from the human demonstration, the final point holds the same as the ground truth. 

\begin{figure*}[!htbp]
    \centering
    \includegraphics[width=1.0\linewidth]{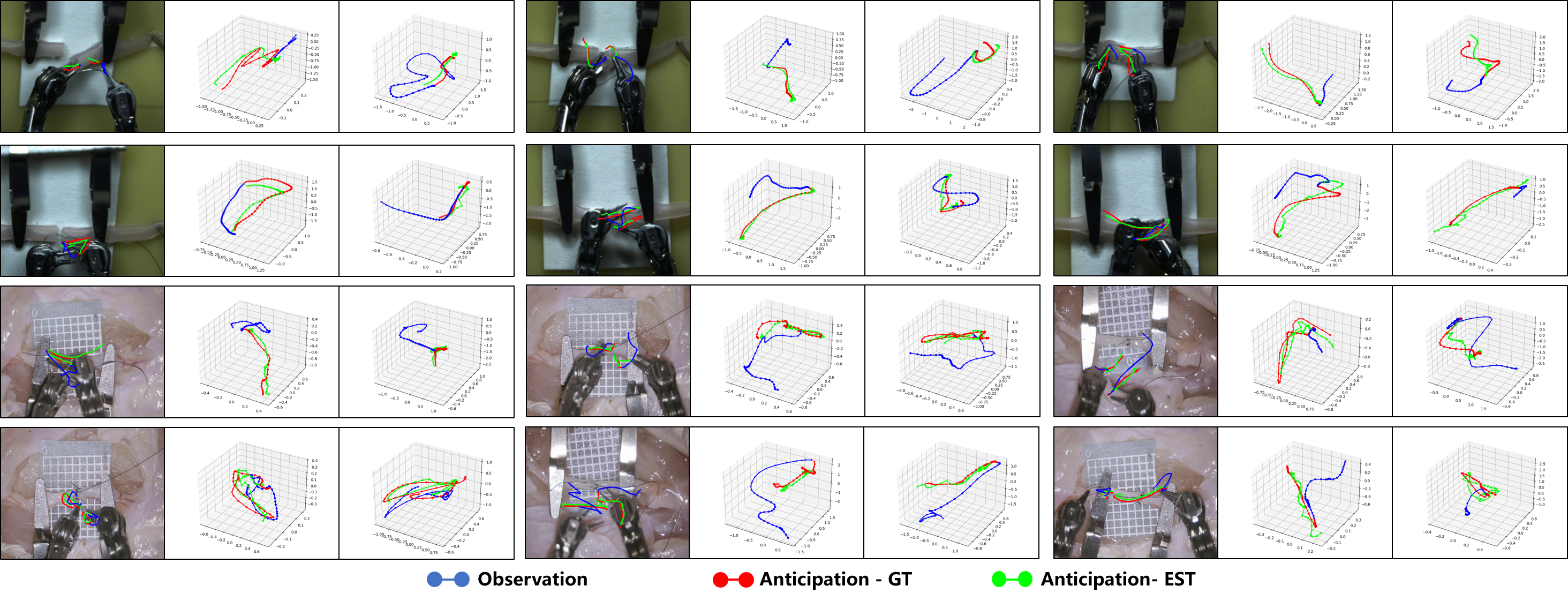}
    \caption{Additional qualitative results of our proposed method on both the phantom dataset and the ex-vivo dataset. The 2D and 3D trajectory anticipations are visualized simultaneously. The blue, red, and green lines indicate the observation, ground truth, and the anticipation of our method, respectively.}
    \label{fig:more_result_phantom_exvivo}
    \vspace{-10pt}
\end{figure*}

Several extensive examples of the combination of 2D and 3D trajectory anticipation by our proposed method are shown in Fig. \ref{fig:more_result_phantom_exvivo}, where the upper part provides examples on the phantom dataset and the lower part shows the results on the ex-vivo dataset with more complex operations. The provided results have demonstrated the superior performance of our method for trajectory anticipation on different surgical scenes and surgical actions.

\begin{table*}[!htbp]
\centering
\scriptsize
\renewcommand\arraystretch{1.5}
\caption{Ablated results of 2D and 3D surgical trajectory anticipation with different network architectures and training schemes}
\begin{tabular}{cccccccccc}
\hline
\multirow{2}{*}{Description}&\multirow{2}{*}{Approaches}& \multicolumn{4}{c}{2D Trajectory} & \multicolumn{4}{c}{3D Trajectory} \\
  & & ADE & FDE & FRD & HD & ADE & FDE & FRD & HD\\
  \hline
  \multirow{5}{*}{Multi-modal Feature Extraction} & w/o semantics and motion $\{z^{\mathrm{M}}, z^{\mathrm{F}}\}$  &  13.64 & 16.16 & 20.13 & 14.85 & 30.71   & 48.54  & 58.19  & 42.55    \\
 & w/o detailed semantics $\{\mathrm{M}^{nd}, \mathrm{M}^{tr}\}$  & 11.28  & 14.60 & 16.65 & 12.69 &  26.92  & 42.36  & 49.49  & 36.08 \\
 & w/o operating semantics $\{\mathrm{M}^{l}, \mathrm{M}^{r}, \mathrm{M}^{tr}\}$   & 11.55& 14.60 & 16.70 & 12.68 &  26.64  & 42.09  & 49.18  & 35.75  \\
 & w/o motion $\mathrm{F}$   & 12.84& 15.22 & 18.78 & 14.28 & 27.95  & 45.08  & 53.61  & 39.67    \\
 & w/o instruction $\mathrm{\Omega}$   & 11.50 & 14.21 & 16.37 & 12.29 & 26.52   & 42.18  & 49.33  & 34.06    \\
\hline
\multirow{3}{*}{Cross-modal Feature Fusion} & w/o refine feature map $z_f$   & 14.44& 14.06 & 22.06 & 17.17 &  73.10  & 77.80  & 112.91  & 80.01   \\
 & w/o long-short term $\{z_{cur},z_{short}\}$   & 12.14 & 14.99 & 17.45 & 12.76 & 27.49   & 43.72  & 50.88  & 37.01   \\
 & w/o learnable gate $\alpha_z$   & 12.02 & 14.11 & 16.77 & 12.12 &  28.07  & 43.30  & 51.06  & 36.72   \\
 \hline
Full Model & Ours   & \textbf{11.20}  & \textbf{13.81}   & \textbf{15.63}  & \textbf{11.28} & \textbf{25.28}   & \textbf{40.83}  & \textbf{46.74} & \textbf{34.13}     \\
\hline
 
\label{tab:ab_visual_input}
\end{tabular}
\end{table*}

\section{Critical Analysis and Discussion}
\subsection{What is the influence of multiple visual representations?}
To evaluate the influence of extracted multi-modal feature maps, we have conducted several ablation studies by removing part of them step by step, and the results are reported in the upper part of Table \ref{tab:ab_visual_input} on the phantom dataset D-I. The first four rows refer to the ablation studies of extracted semantic segmentations and motion maps by our proposed multi-task learning framework. The experiment removing semantic and motion maps is noted as w/o semantics and motion, and the case that only removes the motion maps is denoted as w/o motion. Meanwhile, we separate the categories of semantics into removing detailed objects $\{\mathrm{M}^{nd},\mathrm{M}^{tr}\}$ and removing other operating objects $\{\mathrm{M}^l,\mathrm{M}^r,\mathrm{M}^{tr}\}$, and the difference between them is that the needle and thread are with tiny structure while existing methods fail to precisely recognize them. From the reported quantitative results, it can be found that both semantic feature maps and motion maps contribute a lot to the improvement of both 2D and 3D trajectory anticipation, and the accurate recognition of detailed objects imposes great importance for the instrument-tissue interaction and affordance anticipation, which demonstrates the effectiveness of our designed multi-modal feature extraction network. Compared with the model without semantic and motion maps, our full model achieves a performance improvement of  21.79\%, 17.02\%, 28.79\% and 31.65\% for the metrics ADE, FDE, FRD, and HD on the 2D case, and 21.48\%, 18.89\%, 24.50\% and 24.67\% for the metrics ADE, FDE, FRD, and HD on the 3D case, respectively. In addition, the last row refers to removing the task instructions, and it can be found that the performance decreases, which proves that providing the detailed task description can help the network learn how to generate corresponding trajectories more robustly.

\begin{table}[!htbp]
\centering
\scriptsize
\renewcommand\arraystretch{1.5}
\caption{Ablated results of 2D and 3D surgical trajectory anticipation with different fusion schemes and anticipation periods}
\begin{tabular}{ccccc}
\hline
\multirow{2}{*}{Approaches}& \multicolumn{2}{c}{2D Trajectory} & \multicolumn{2}{c}{3D Trajectory}     \\
  & ADE & FDE & ADE & FDE \\
  \hline

$((z^{\mathrm{I}} \odot z^{\mathrm{\Omega}})\oplus z^{\mathrm{P/J}}) \odot (z^{\mathrm{M}} \oplus z^{\mathrm{F}})$   & \textbf{11.20}   & \textbf{13.81}   &  \textbf{25.28}    &  \textbf{40.83}      \\
$((z^{\mathrm{I}} \odot z^{\mathrm{\Omega}})\odot z^{\mathrm{P/J}}) \odot (z^{\mathrm{M}} \odot z^{\mathrm{F}})$  & 11.77   & 14.44   & 27.17     & 43.40        \\
$((z^{\mathrm{I}} \oplus z^{\mathrm{\Omega}})\oplus z^{\mathrm{P/J}}) \oplus (z^{\mathrm{M}} \oplus z^{\mathrm{F}})$  &  11.47     &  13.90     & 28.23     & 44.83  \\
\hline
$\{\mathrm{T_{obs}},\mathrm{T_{pred}}\}=\{5s,3s\}$   &  9.73  & 12.09   & 18.76     & 31.54     \\
$\{\mathrm{T_{obs}},\mathrm{T_{pred}}\}=\{5s,5s\}$   &  11.20   & 13.81    &  25.28     &   40.83    \\
$\{\mathrm{T_{obs}},\mathrm{T_{pred}}\}=\{5s,7s\}$    & 12.33      & 14.82      &32.26    &48.31      \\
$\{\mathrm{T_{obs}},\mathrm{T_{pred}}\}=\{5s,10s\}$  & 13.36   & 15.45   & 37.84     & 53.55        \\
$\{\mathrm{T_{obs}},\mathrm{T_{pred}}\}=\{5s,15s\}$    &  14.88      & 16.01      & 45.24     &  57.21    \\
\hline
 
\label{tab:ab_progressive}
\end{tabular}
\end{table}

\subsection{How does the cross-modal fusion improve the performance?}
In the second part of Table \ref{tab:ab_visual_input}, quantitative results of ablation studies for cross-modal feature fusion are reported on the phantom dataset D-I. The first row refers to the experiment of removing all feature fusion block with only the coarse-level feature map as input for the trajectory prediction noted as w/o refine feature map. It can be noted that the performance decreases a lot without our proposed fusion module. The second row refers to removing our proposed long-short-term fusion mechanism with only $z_{long}$ as input, which indicates the same attention on all timesteps. With our well-designed long-short-term feature aggregation, the performance improves about 8.39\%, 8.54\%, 11.64\%, and 13.12\% for the metrics ADE, FDE, FRD, and HD on the 2D case, and 8.74\%, 7.08\%, 8.86\%, and 8.44\% for the metrics ADE, FDE, FRD, and HD on the 3D case. The last row refers to removing the process of dynamic weighted sum of three feature maps, and we apply the same weights for the three terms instead. It can be found that by introducing a learnable gate contributes to a better performance for both 2D and 3D cases compared with an invariant constant.

Different cross-modal feature fusion schemes are also evaluated in the upper part of Table \ref{tab:ab_progressive}. Our proposed method follows  Equ. (\ref{feature_concat}) to compute the fused feature maps based on the observation, and two more types of fusion methods are implemented as ablation. From the reported results, it can be found that our designed fusion mechanism achieves the best performance compared to the other two schemes on the dataset D-I, especially for the 3D trajectory prediction. Our method achieves a performance improvement of 2.41\% and 0.65\% for ADE and FDE on the 2D case, respectively, and 7.48\% and 6.29\%  on the 3D case compared with the second-best scheme.

\begin{figure}[!htbp]
    \centering
    \includegraphics[width=1.0\linewidth]{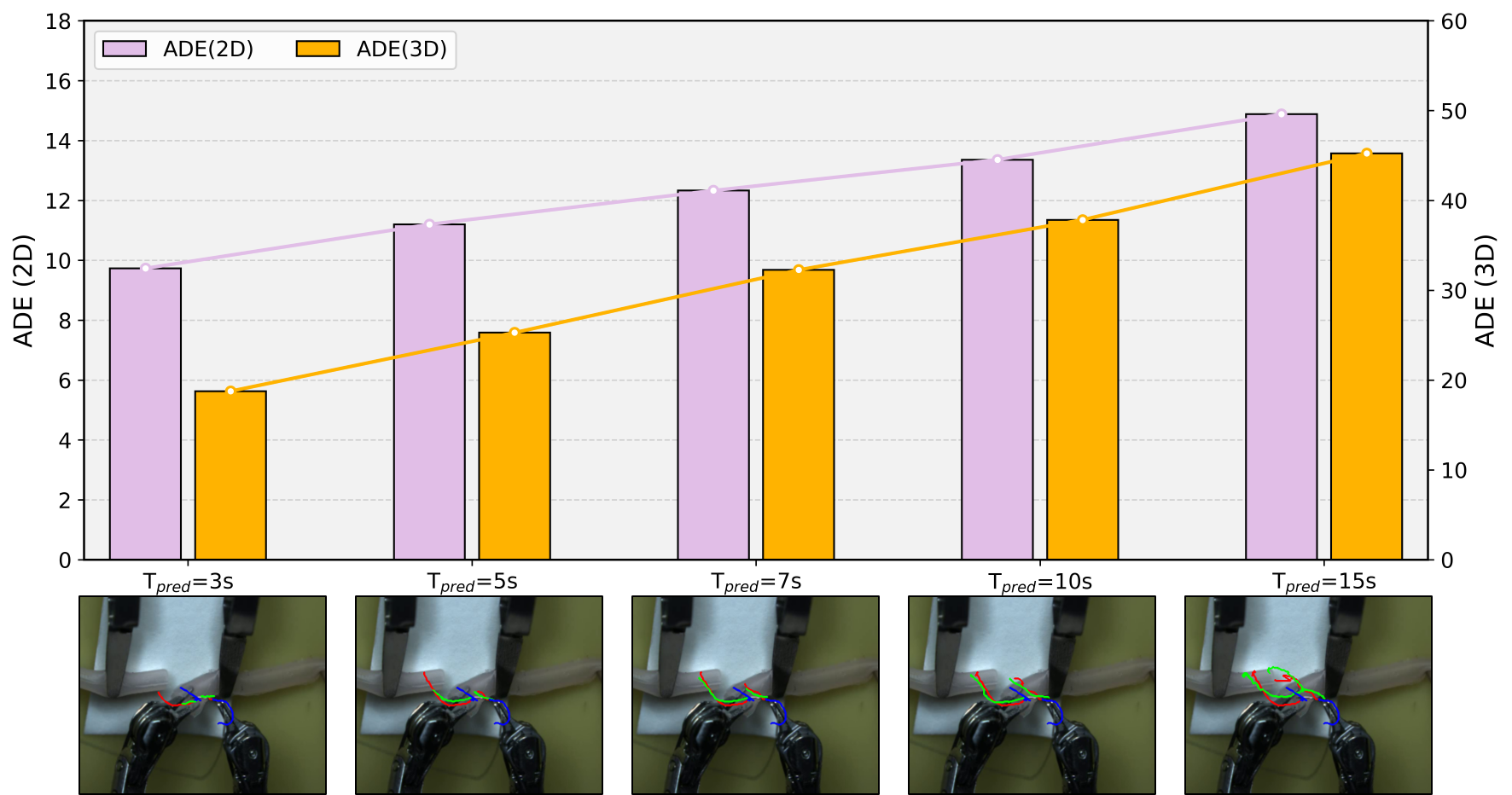}
    \caption{Performance of 2D and 3D trajectory anticipation for different anticipation periods on the phantom case. The left axis refers to the error value of the 2D trajectory, and the right axis refers to the error value of the 3D trajectory. The lower part shows the predicted 2D trajectories with the same observation.}
    \label{fig:vis_seq_len}
    \vspace{-10pt}
\end{figure}

\begin{figure}[!htbp]
    \centering
    \includegraphics[width=1.0\linewidth]{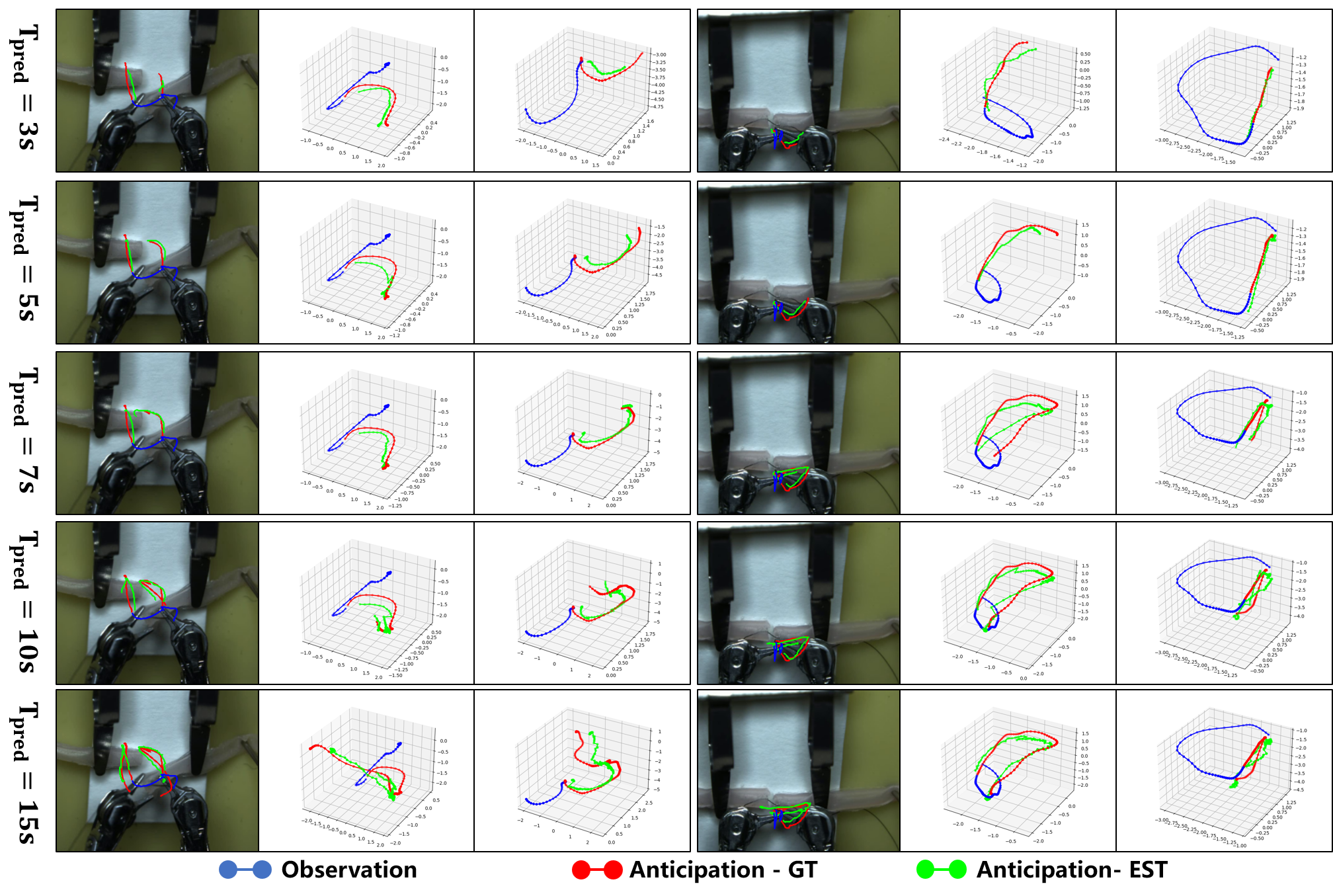}
    \caption{Qualitative results of our proposed method for different anticipation periods $\mathrm{T_{pred}=\{3s,5s,7s,10s,15s\}}$. The 2D and 3D trajectory anticipations are visualized simultaneously. The blue, red, and green lines indicate the observation, ground truth, and the anticipation of our method, respectively.}
    \label{fig:vis_pred_seq}
    \vspace{-10pt}
\end{figure}

\subsection{What is the influence of the anticipation period of the surgical trajectory?}
For the results mentioned above, the observation period is the past five seconds, and the prediction period is the next five seconds by default. By changing the prediction period, we can evaluate the accuracy and generalization ability of our method for long-term trajectory anticipation. In the lower part of Table \ref{tab:ab_progressive}, experiments are conducted to extend the prediction period $\mathrm{T_{pred}}$ from three seconds to fifteen seconds on the dataset D-I, where five seconds is the default experimental setting for other experiments. We have also provided a more intuitive bar plot in Fig. \ref{fig:vis_seq_len} and several visualization results of different prediction periods in Fig. \ref{fig:vis_pred_seq}, where the network takes the same observation in the past five seconds. It can be found that for the short period of five seconds, our method generates the trajectories with a very similar shape to the human demonstration. When the period extends to ten seconds, our proposed framework generalizes well to relatively long-term prediction on both 2D and 3D cases. For the case of anticipating future fifteen seconds, our method can still provide a rough trend of movements that guides the surgical instruments towards the target position. Those facts demonstrate that our proposed framework has the potential to be applied for long-term surgical anticipation.

\begin{table}[!htbp]
\centering
\scriptsize
\renewcommand\arraystretch{1.5}
\caption{Ablated results of 2D and 3D surgical trajectory anticipation with different coefficients $\alpha_{pos}^{3D}$}
\begin{tabular}{ccccccccc}
\hline
\multirow{2}{*}{$\alpha_{pos}^{3D}$}& \multicolumn{4}{c}{2D Trajectory} & \multicolumn{4}{c}{3D Trajectory}     \\
  & ADE & FDE & FRD & HD& ADE & FDE& FRD & HD \\
  \hline
15.0 & 12.15  &  14.26    &  17.23   & 12.78 &  27.51 &  44.14    & 51.74     & 37.38  \\
10.0  &  \underline{11.39}  &  \underline{13.95}     & \underline{16.10}      & \underline{11.56}   & 26.78  &  42.32     & 49.36      &36.24    \\
5.0    & \textbf{11.20} & \textbf{13.81}     & \textbf{15.63}     & \textbf{11.28}  & \textbf{25.28}  & \textbf{40.83}      & \textbf{46.74}      & \textbf{34.13}   \\
1.0 &  11.80 &  14.16     &  16.31   &  11.63  & \underline{25.64}  &  \underline{41.66}     & \underline{48.09}      &    \underline{34.32}  \\
0.5 & 12.39  &  14.20     &  17.13    & 12.26 &  27.04 &  42.50     &  49.80     & 34.82  \\
0.1 & 13.23  &  14.54     &  19.00    & 12.97 &  28.08 &  44.86     &  52.60     & 35.69  \\

\hline
 
\label{tab:ab_coefficients}
\end{tabular}
\end{table}

\begin{figure}[!htbp]
    \centering
    \includegraphics[width=1.0\linewidth]{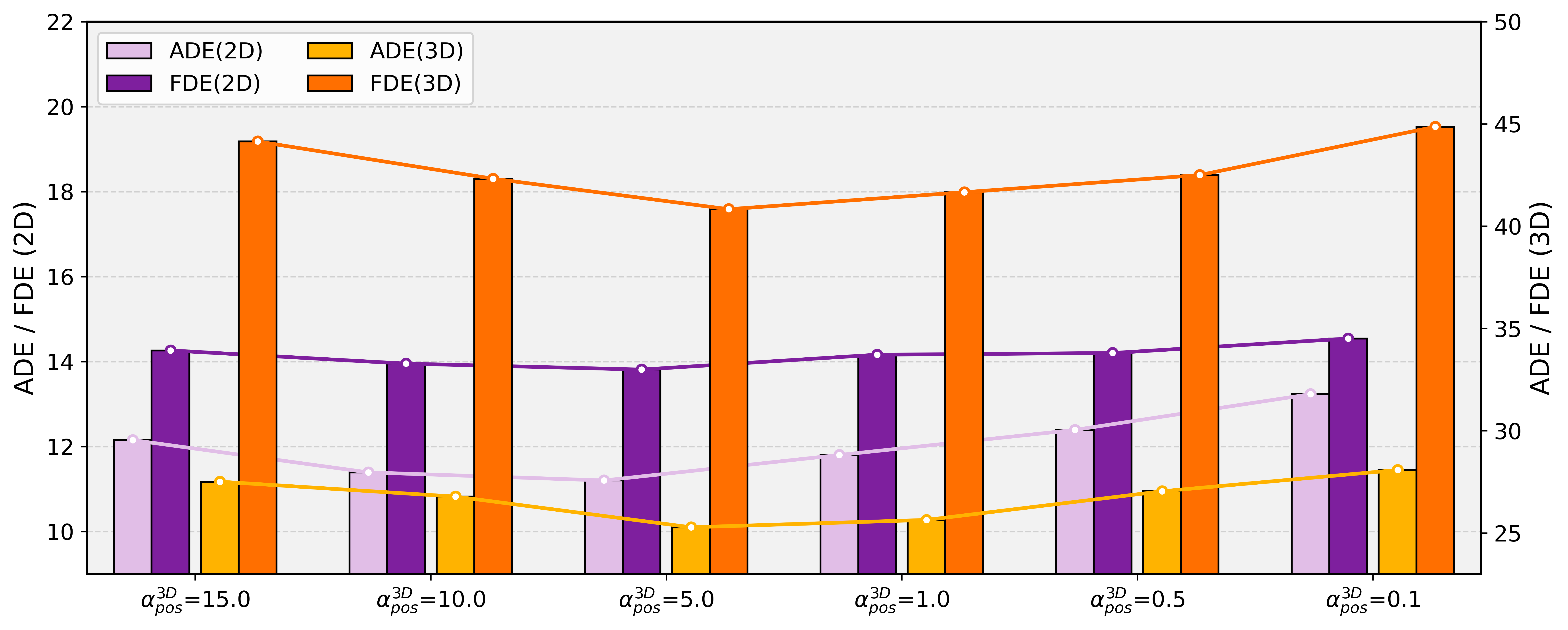}
    \caption{Performance of 2D and 3D trajectory anticipation for different choices of the coefficient $\alpha_{pos}^{3D}$. The left axis refers to the error value of the 2D trajectory, and the right axis refers to the error value of the 3D trajectory.}
    \label{fig:balance_coefficient}
    \vspace{-10pt}
\end{figure}

\begin{figure*}[!htbp]
    \centering
    \includegraphics[width=1.0\linewidth]{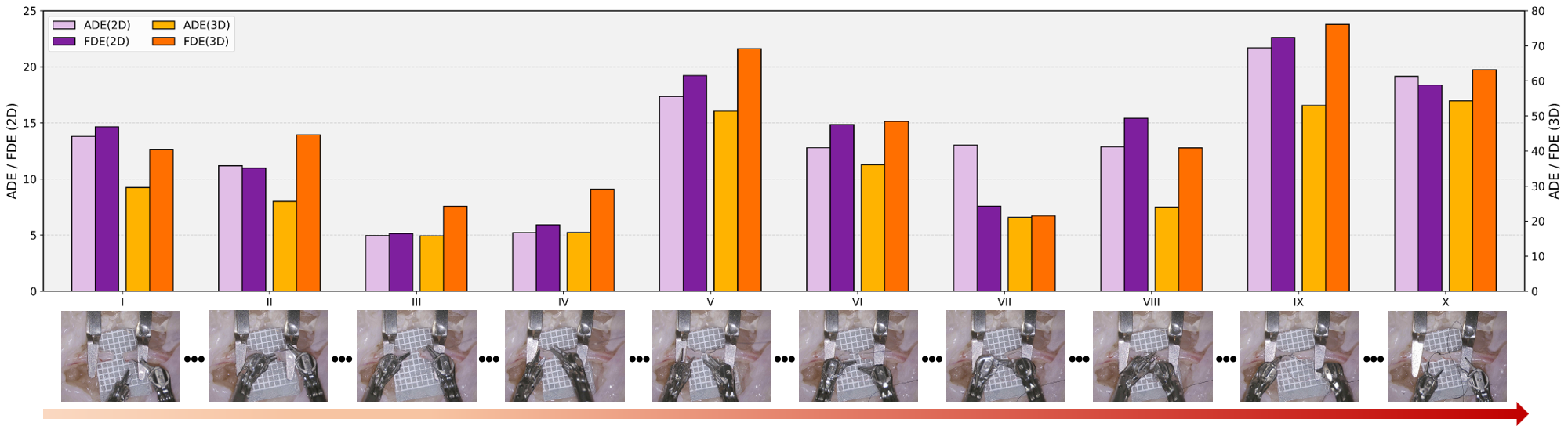}
    \caption{Performance of 2D and 3D trajectory anticipation for different surgical actions on the ex-vivo case. The left axis refers to the error value of the 2D trajectory, and the right axis refers to the error value of the 3D trajectory. The lower part shows the representative frames for each action.}
    \label{fig:vis_act_specific}
    \vspace{-10pt}
\end{figure*}

\subsection{How to balance the performance between 2D and 3D anticipation?}
During the training stage of our proposed framework, the balance of 2D and 3D trajectory prediction is very important and the coefficient $\alpha_{pos}^{3D}$ plays a crucial role in the training constraint $\mathcal{L}_{pos}$. The results of parameter ablation studies of $\alpha_{pos}^{3D}$ are reported in Table \ref{tab:ab_coefficients} and the corresponding bar plot is provided in Fig. \ref{fig:balance_coefficient}. For a fair comparison, all the experiments are conducted on the phantom dataset with the same training epochs based on the same extracted visual representations. It can be found that our choice is the best compared with other coefficients and achieves an improvement of 1.70\% and 1.01\% for ADE and FDE on the 2D case, and 1.42\% and 2.03\%  on the 3D case with respect to the second-best choice.  More importantly, the performance of 3D trajectory anticipation is more sensitive to the choice of $\alpha_{pos}^{3D}$, which mainly results from the data quality of derived kinematics and the complexity of prediction in 3D space. From the provided parameter ablations, we can conclude that our choice balances the tasks of 2D and 3D trajectory prediction well and contributes to the best performance.

\subsection{What is the difference in the anticipation performance among different surgical actions?}
In Table \ref{tab:ab_phase}, we report the results for each action on our proposed ex-vivo dataset, and there are ten predefined actions in total. In Fig. \ref{fig:vis_act_specific}, we have also provided a bar plot to visualize the performance. It can be found that the performance varies a lot among different surgical actions, which results from the complexity and the operating time for each process. For actions III and IV, the movements are simple and frequent, resulting in a better performance compared to the average. For the action I of needle picking and preparation, and the actions V and IX for conducting and tying the knot, the performance decreases below the average performance, which indicates that those actions are much more difficult and challenging. We have also found that the distribution between the phantom and ex-vivo datasets also varies a lot, and the reasons can be concluded from two aspects. On the one hand, the surgical scenes for ex-vivo anastomosis are more complex and similar to real-world clinical settings, and the trajectory distribution across actions is different from that of the phantom scenarios. On the other hand, the time consumed for each action on ex-vivo case is much longer than that on the phantom case, which can also result in different error distributions shown in the table. Those results also provide future directions to focus on the complex processes like knotting, and attempt to improve the performance on these challenging actions.

\begin{table}[!htbp]
\centering
\scriptsize
\setlength\tabcolsep{3pt}
\renewcommand\arraystretch{1.5}
\caption{Results of 2D and 3D surgical trajectory anticipation for different surgical actions}
\begin{tabular}{ ccccccccccc}
\hline
  
 Action & I & II & III &IV & V& VI & VII & VIII & IX & X \\
  \hline
  ADE-2D & 13.80  & 11.18  & 4.95 & 5.22 &17.36 &  12.78&  13.01 & 12.87  &21.69   &19.14  \\
  FDE-2D  & 14.66  & 10.97  & 5.14 & 5.92 & 19.22& 14.85 & 7.58  & 15.41  & 22.63  & 18.38 \\
  ADE-3D  & 29.57  &  25.61 & 15.77 & 16.74 & 51.37& 36.03 & 21.04  & 24.02  & 53.01  & 54.28 \\
  FDE-3D  & 40.47  & 44.60  & 24.22 & 29.13 & 69.17& 48.45 & 21.50  & 40.87  & 76.09  & 63.19 \\

\hline
 
\label{tab:ab_phase}
\vspace{-10pt}
\end{tabular}
\end{table}

 
\subsection{Limitation and Future Work}
Our proposed framework has achieved remarkable progress in both 2D and 3D trajectory anticipation by implementing the well-designed multi-modal feature extractor and proposed cross-modal feature fusion scheme. However, there still exist limitations of our current work, and future studies will attempt to solve them and improve the performance. First, our proposed method mainly focuses on microsurgical anastomosis, and all experiments are conducted based on our collected data from our established robotic platform. An extensive evaluation of other typical surgeries is needed to prove the strong generalization ability across different surgeries. Second, a two-stage training strategy is implemented to train the two modules separately, and our future work will attempt to combine them in an end-to-end manner. Finally, the current experimental evaluation is mainly based on our offline dataset, and we will try to incorporate the proposed model into our surgical platform to provide visual guidance for both copilot assistance and surgery autonomy. In addition, our future work will focus on introducing uncertainty to our trajectory anticipation as motion priors for surgical video generation, and applying the framework for a unified surgical world model on microsurgical anastomosis.


 \section{Conclusions}
 In this work, we propose to anticipate long-term 2D and 3D surgical trajectories for microsurgical anastomosis, which contains two critical networks, i.e., a multi-modal feature extraction network and a cross-modal feature fusion network. Our proposed multi-modal feature extraction network extracts diverse feature representations from both semantic and temporal domains in a semi-supervised multi-task manner, and the derived semantic maps and motion fields preserve the inherent relationships of different surgical objects in instrument-tissue interactions. The cross-modal feature fusion network incorporates a coarse-to-fine fusion scheme and a long-short-term fusion block to dynamically alleviate different weights for temporal dependencies. Based on the extracted multi-modal feature maps, the refined fused feature is derived, and trajectory anticipation is conducted. Detailed experiments have been conducted on our proposed datasets, including both the phantom and ex-vivo scenarios, which are collected on our established surgical robotic platform for microsurgical anastomosis. Quantitative and qualitative results have demonstrated that our proposed framework achieves superior performance compared with existing methods, and the design of different blocks contributes to the improved accuracy and robustness across different surgical scenes.


\bibliographystyle{IEEEtran}
\bibliography{Reference}


\end{document}